\documentclass[preprint,12pt]{elsarticle}
\usepackage{geometry}
\usepackage{graphicx}
\usepackage{amssymb}
\usepackage{caption}
\usepackage{subcaption}
\usepackage{xcolor}

\journal{Journal Name}

\begin{document}

\begin{frontmatter}


\title{Closed-Loop Adaptation for Weakly-Supervised Semantic Segmentation}



\author{
Zhengqiang Zhang$^1$ ～~
Shujian Yu$^2$ ~~
Shi Yin$^1$ ~~
Qinmu Peng$^1$ ~~
Xinge You$^1$}

\address{
$^1$School of Electronic Information and Communications, Huazhong University of Science and Technology\\
$^2$Department of Electrical and Computer Engineering, University of Florida\\
}

\begin{abstract}
Weakly-supervised semantic segmentation aims to assign each pixel a semantic category under weak supervisions, such as image-level tags. Most of existing weakly-supervised semantic segmentation methods do not use any feedback from segmentation output and can be considered as open-loop systems. They are prone to accumulated errors because of the static seeds and the sensitive structure information. In this paper, we propose a generic self-adaptation mechanism for existing weakly-supervised semantic segmentation methods by introducing two feedback chains, thus constituting a closed-loop system. Specifically, the first chain iteratively produces dynamic seeds by incorporating cross-image structure information, whereas the second chain further expands seed regions by a customized random walk process to reconcile inner-image structure information characterized by superpixels. Experiments on PASCAL VOC 2012 suggest that our network outperforms state-of-the-art methods with significantly less computational and memory burden.
\end{abstract}


\end{frontmatter}


\section{Introduction}
Semantic segmentation of an image refers to the task of assigning each pixel a categorical label, e.g., motorcycle or person~\cite{zhao2017pyramid}. Owing to the rapid development of deep learning, tremendous progress has been made for fully-annotated images. Notable examples include the DeepLab~\cite{chen2018deeplab:} and the PSPNet~\cite{zhao2017pyramid}. These methods assume that the pixel-level labels are available immediately upon request. However, this assumption is over-optimistic, since it could involve the annotation of data by expensive means in terms of cost and labor
time. As a result, weakly-supervised semantic segmentation which only requires a few weak labels, such as bounding box \cite{dai2015boxsup:}, scribble \cite{lin2016scribblesup:}, points \cite{bearman2016what} and tags \cite{long2015fully}, has attracted increasing attention.

This paper focuses on weakly-supervised semantic segmentation with image-level tags. Under the circumstance, the lack of structure information on the organization of image pixels prevents network learning from labels directly, which makes the problem of weakly-supervised semantic segmentation ill-conditioned. Consequently, how to recover the cross-image structure information and the inner-image structure information becomes a pivotal issue. The cross-image structure information describes how to organize pixels for a specific semantic category across multiple images. For example, given this kind of information for the person category, we can cluster pixels (in multiple images) into local regions (e.g., heads and faces) that are discriminative to recognize a person. Unlike cross-image structure information, the inner-image structure information tells us how to organize pixels (in a single image) based on their low-level features like textures or colors. This kind of information always contains details about object boundaries.

\begin{figure}
 
\centering
	\begin{subfigure}[b]{0.90\textwidth}
		 \begin{subfigure}[b]{0.5\textwidth}
			 \centering
			 \begin{subfigure}[b]{0.45\textwidth}
				 \centering
				 \includegraphics[width=\textwidth]{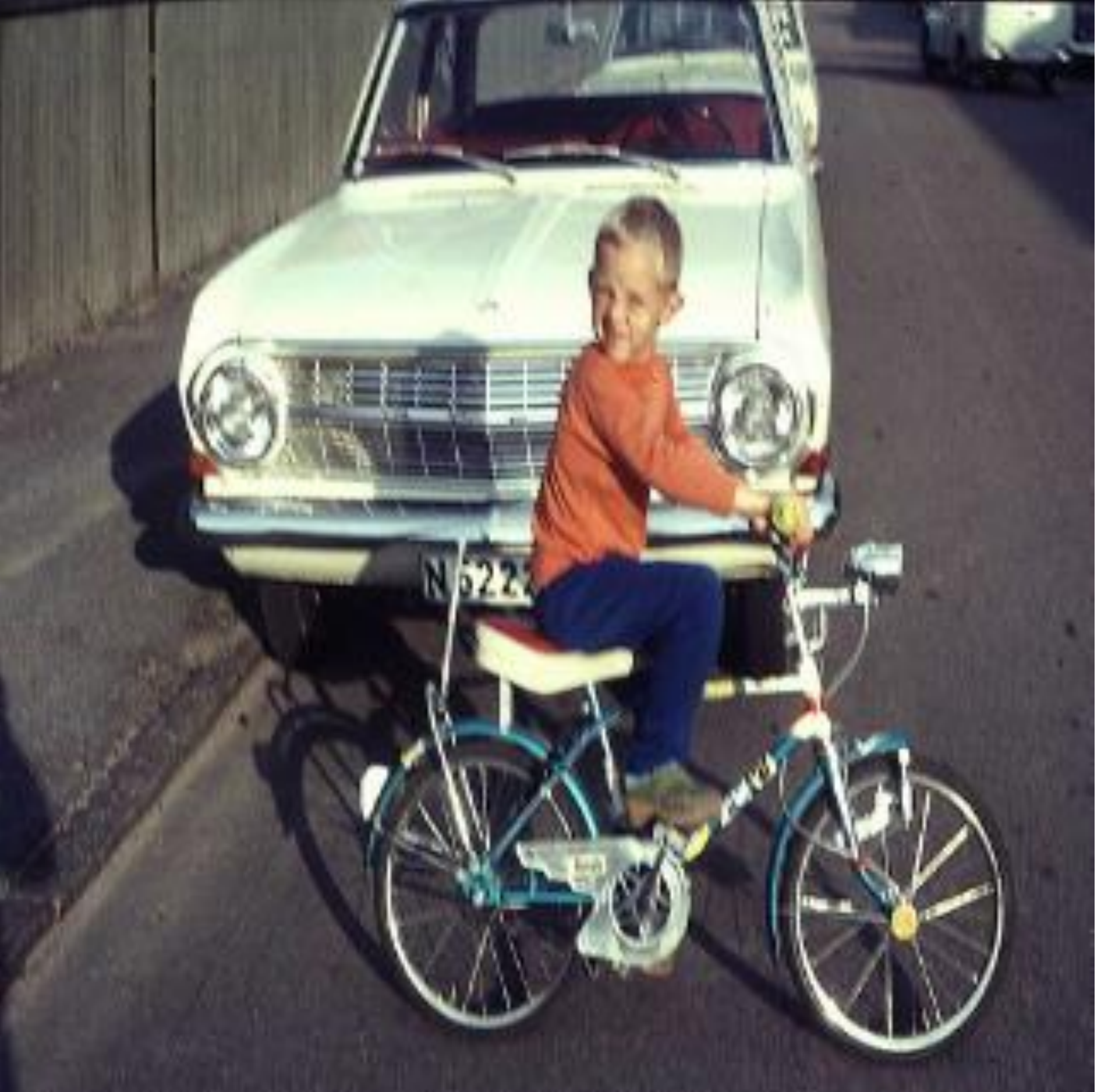}
				 \small{Input}
			 \end{subfigure}
			 \hfill
			 \begin{subfigure}[b]{0.45\textwidth}
				 \centering
				 \includegraphics[width=\textwidth]{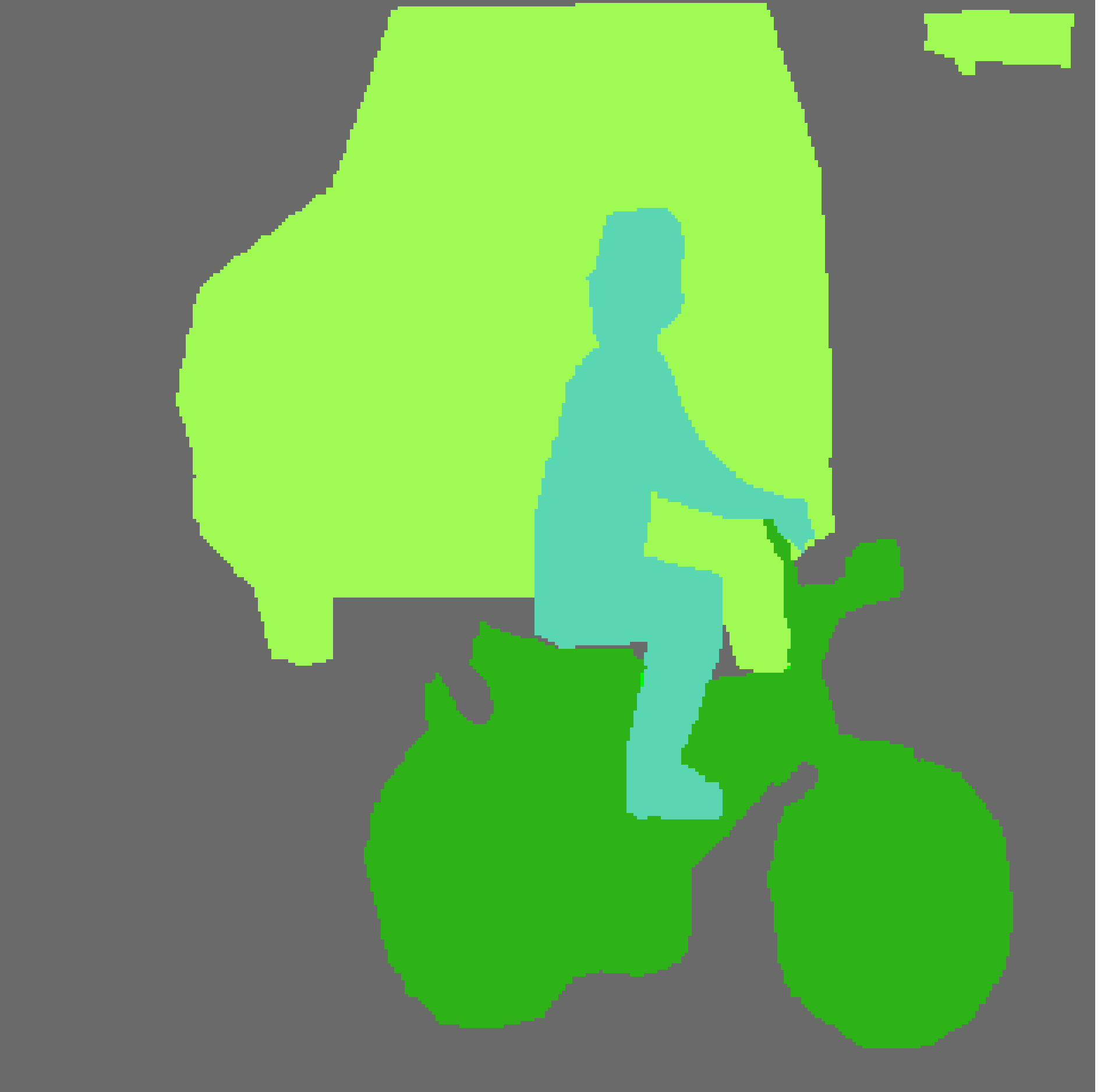}
				 \small{Ground Truth}
			 \end{subfigure}
			 \caption{\small{Image example}}
			 \label{fig:first_figure_a}
		 \end{subfigure}
		 \hspace{0.05\textwidth}
		 \begin{subfigure}[b]{0.40\textwidth}
			 \centering
			 \includegraphics[width=\textwidth]{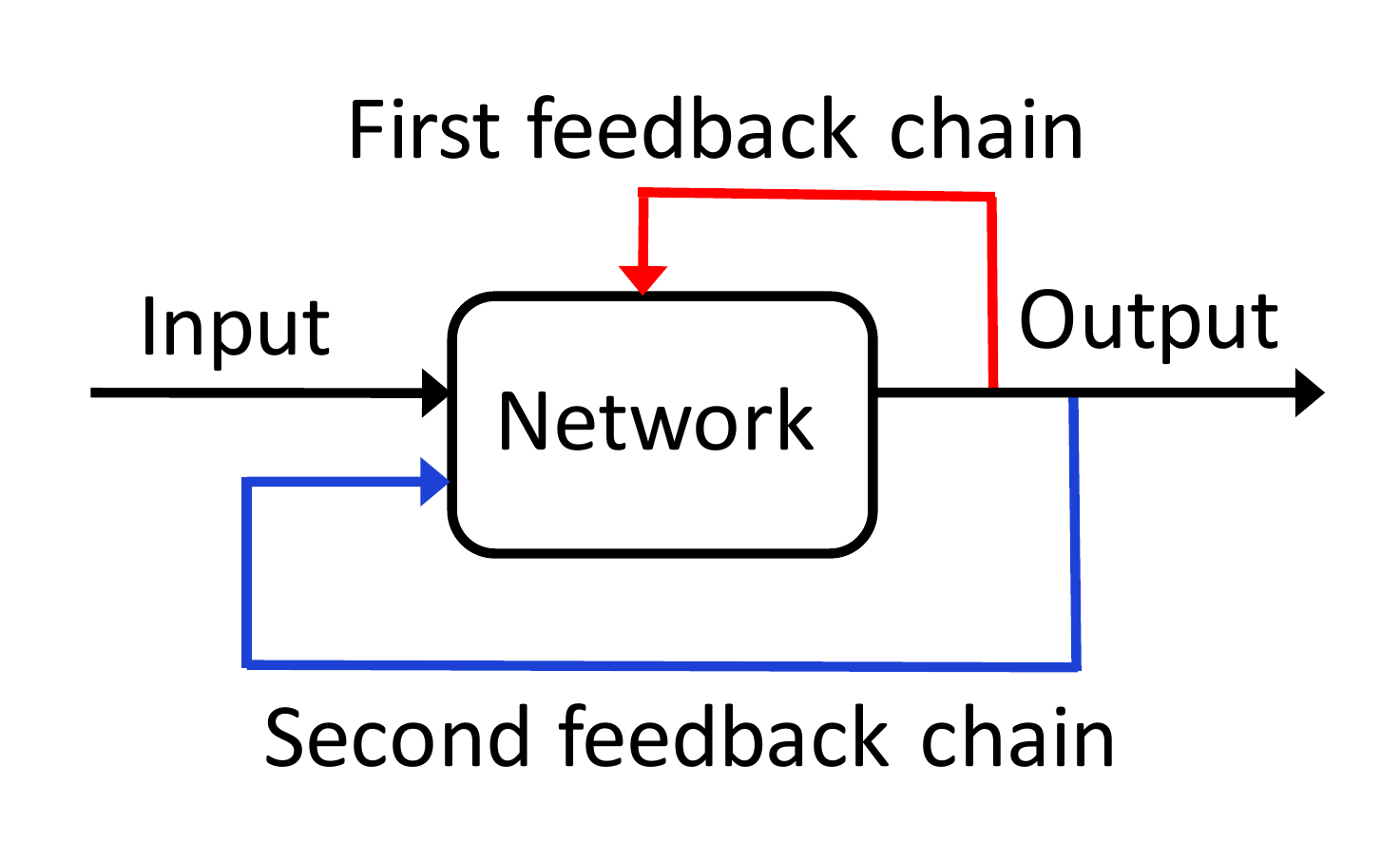}
			
			 \caption{\small{Closed-loop mechanism}}
			 \label{fig:first_figure_framework}
		 \end{subfigure}
	\end{subfigure}

	\begin{subfigure}[b]{0.90\textwidth}
     \begin{subfigure}[b]{0.45\textwidth}
		 \centering
		 \begin{subfigure}[b]{0.3\textwidth}
			 \centering
			 \includegraphics[width=\textwidth]{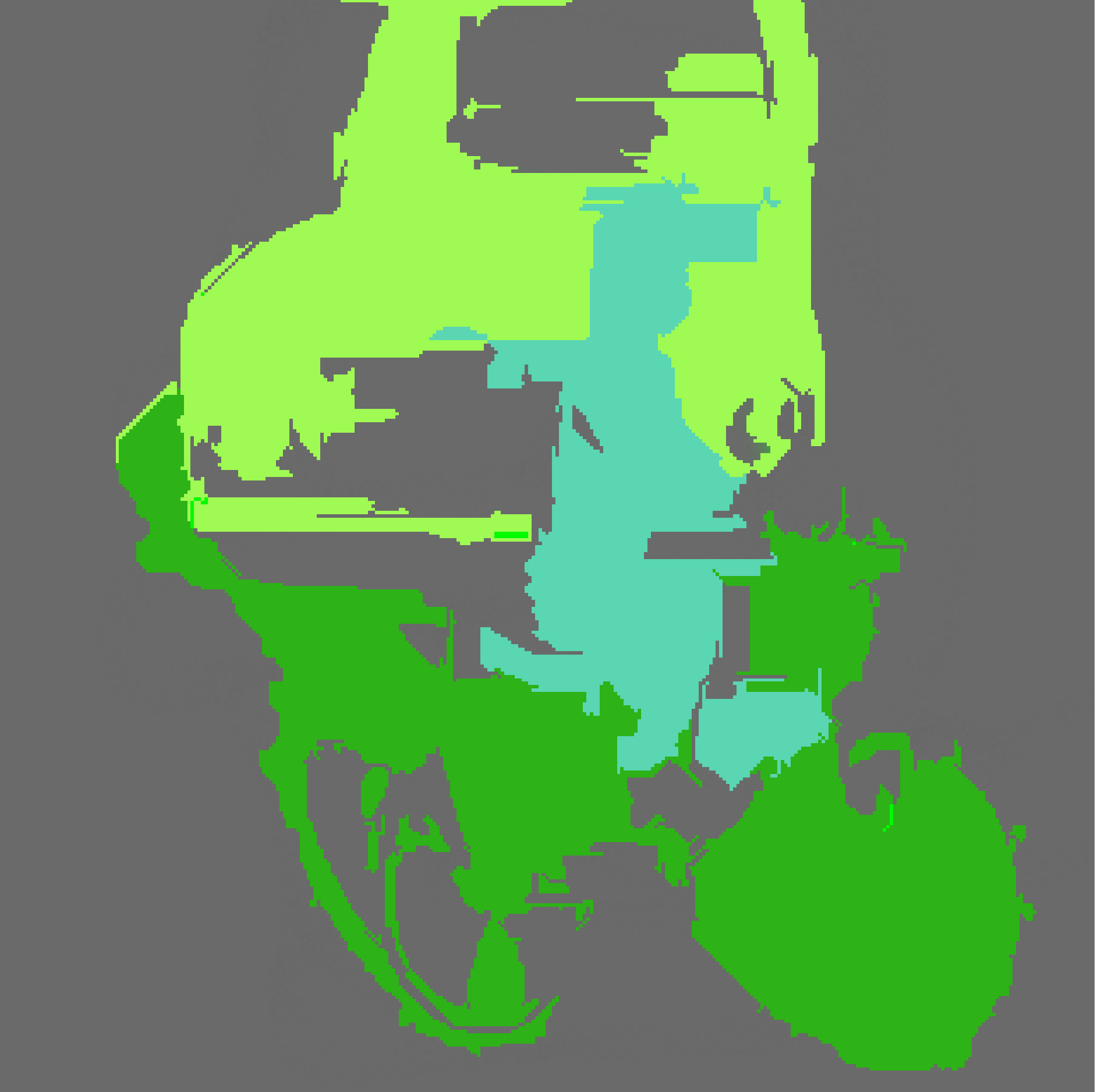}
			 \scriptsize{Epoch \#0}
		 \end{subfigure}
		 \hfill
		 \begin{subfigure}[b]{0.30\textwidth}
			 \centering
			 \includegraphics[width=\textwidth]{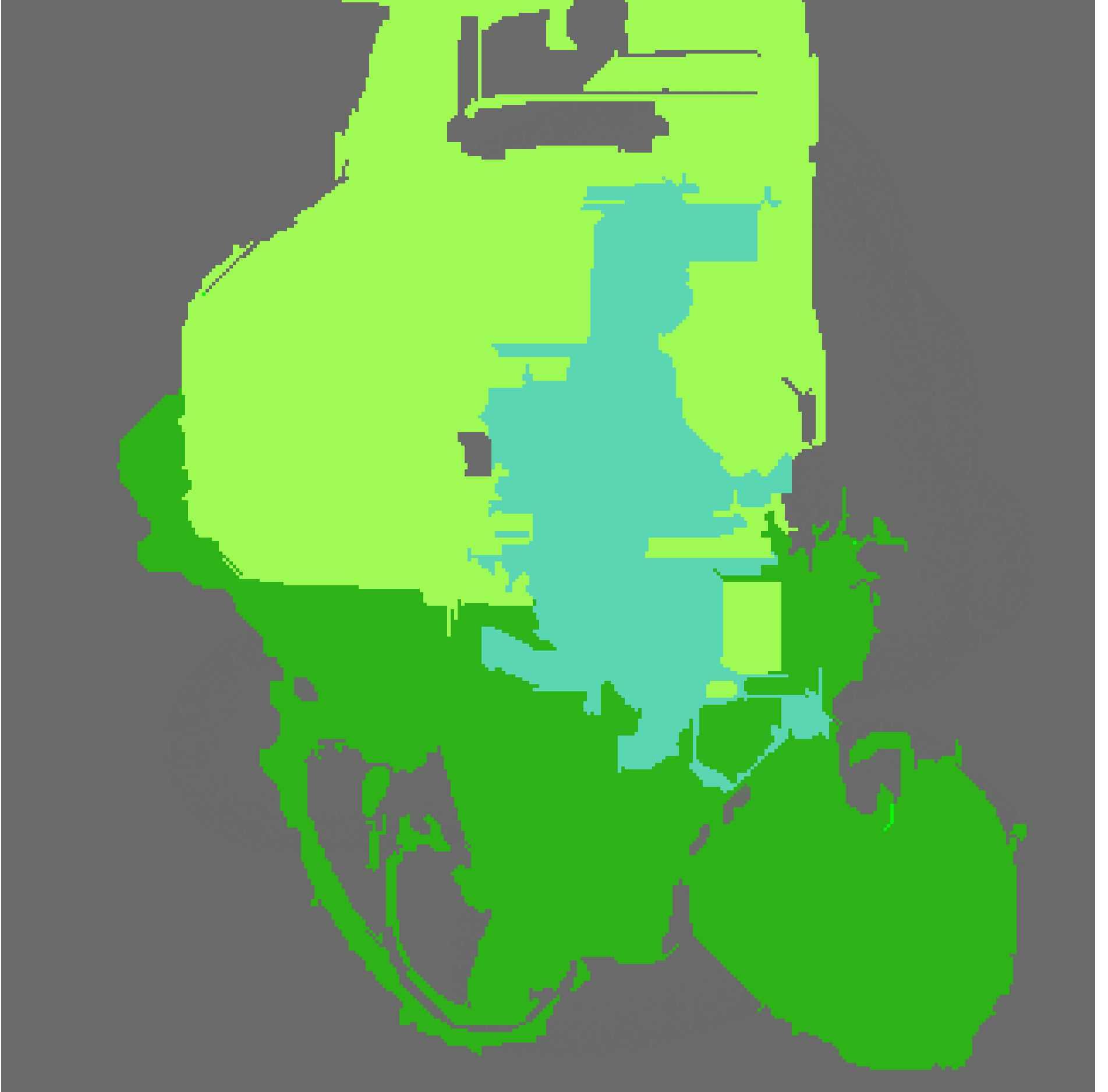}
			 \scriptsize{Epoch \#10}
		 \end{subfigure} 
		 \hfill
		 \begin{subfigure}[b]{0.30\textwidth}
			 \centering
			 \includegraphics[width=\textwidth]{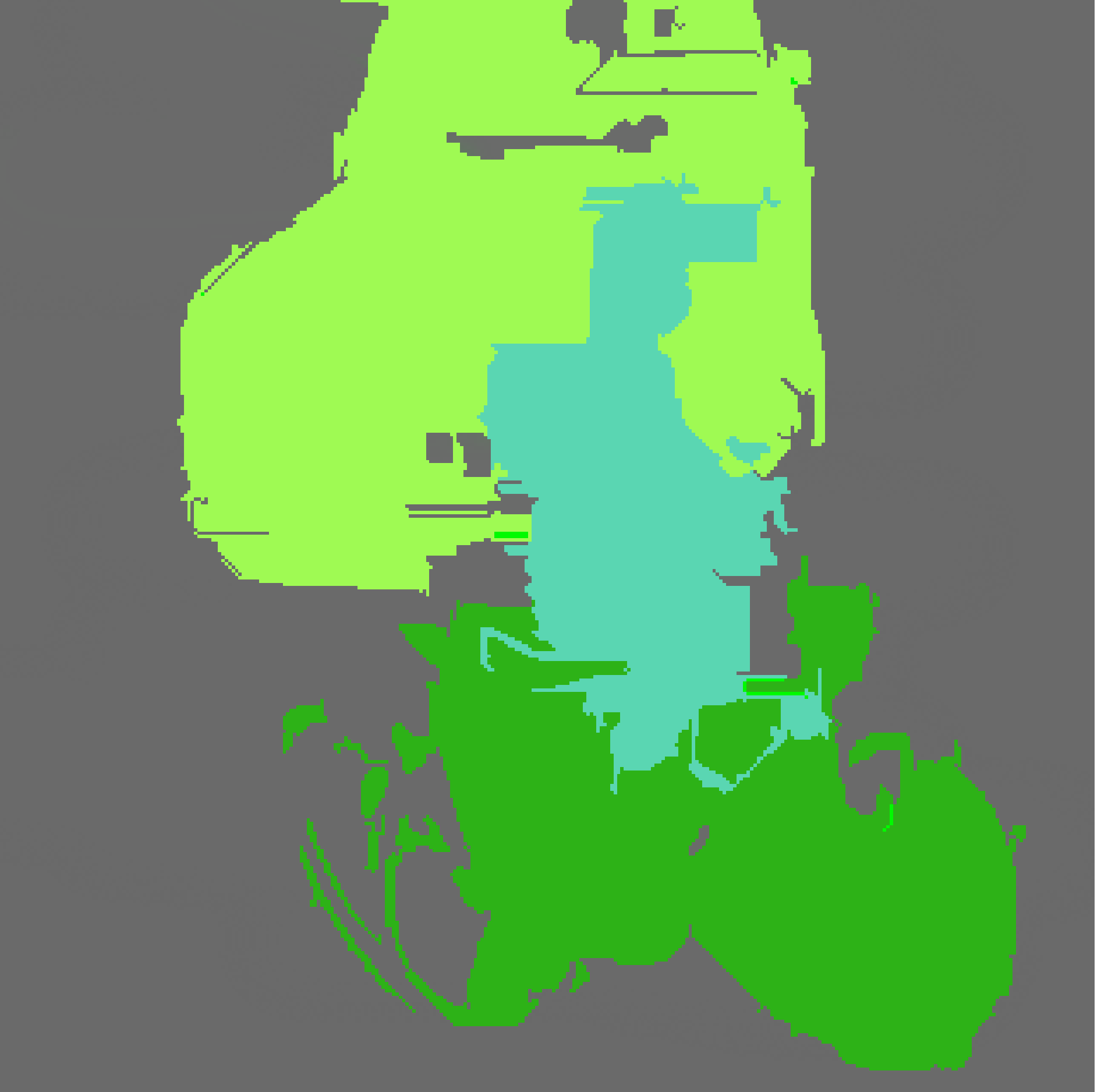}
			 \scriptsize{Epoch \#20}
		 \end{subfigure}
		 \caption{\small{The first feedback chain}}
		 \label{fig:first_figure_feedback_one}
     \end{subfigure}
     \hspace{0.05\textwidth}
     \begin{subfigure}[b]{0.45\textwidth}
     \centering
     \begin{subfigure}[b]{0.3\textwidth}
     \centering
     \includegraphics[width=\textwidth]{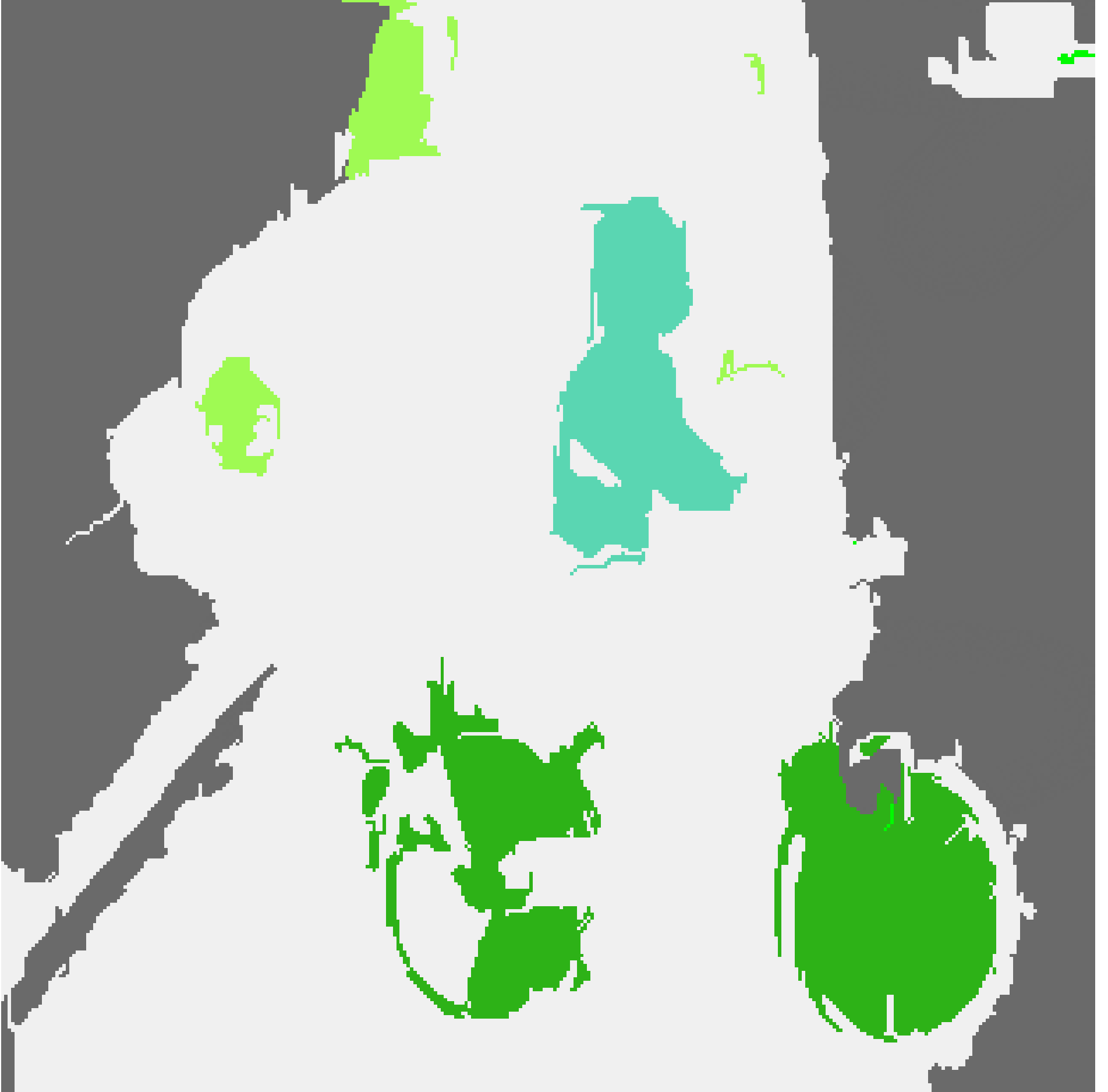}
     \scalebox{0.6}{Input Seed}
     \end{subfigure}
     \hfill
     \begin{subfigure}[b]{0.30\textwidth}
     \centering
     \includegraphics[width=\textwidth]{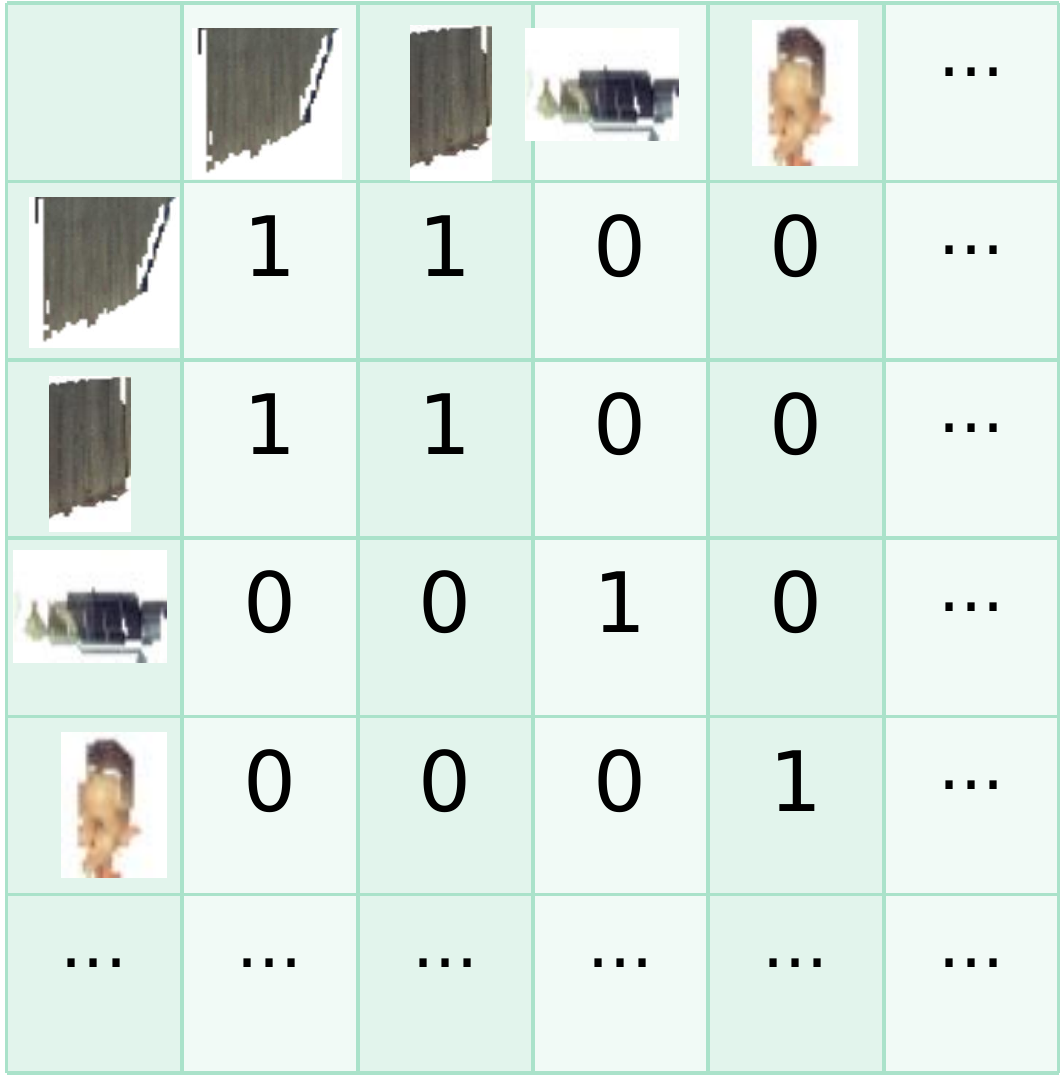}
     \scalebox{0.5}{Relationship Matrix}
     \end{subfigure} 
     \hfill
     \begin{subfigure}[b]{0.3\textwidth}
     \centering
     \includegraphics[width=\textwidth]{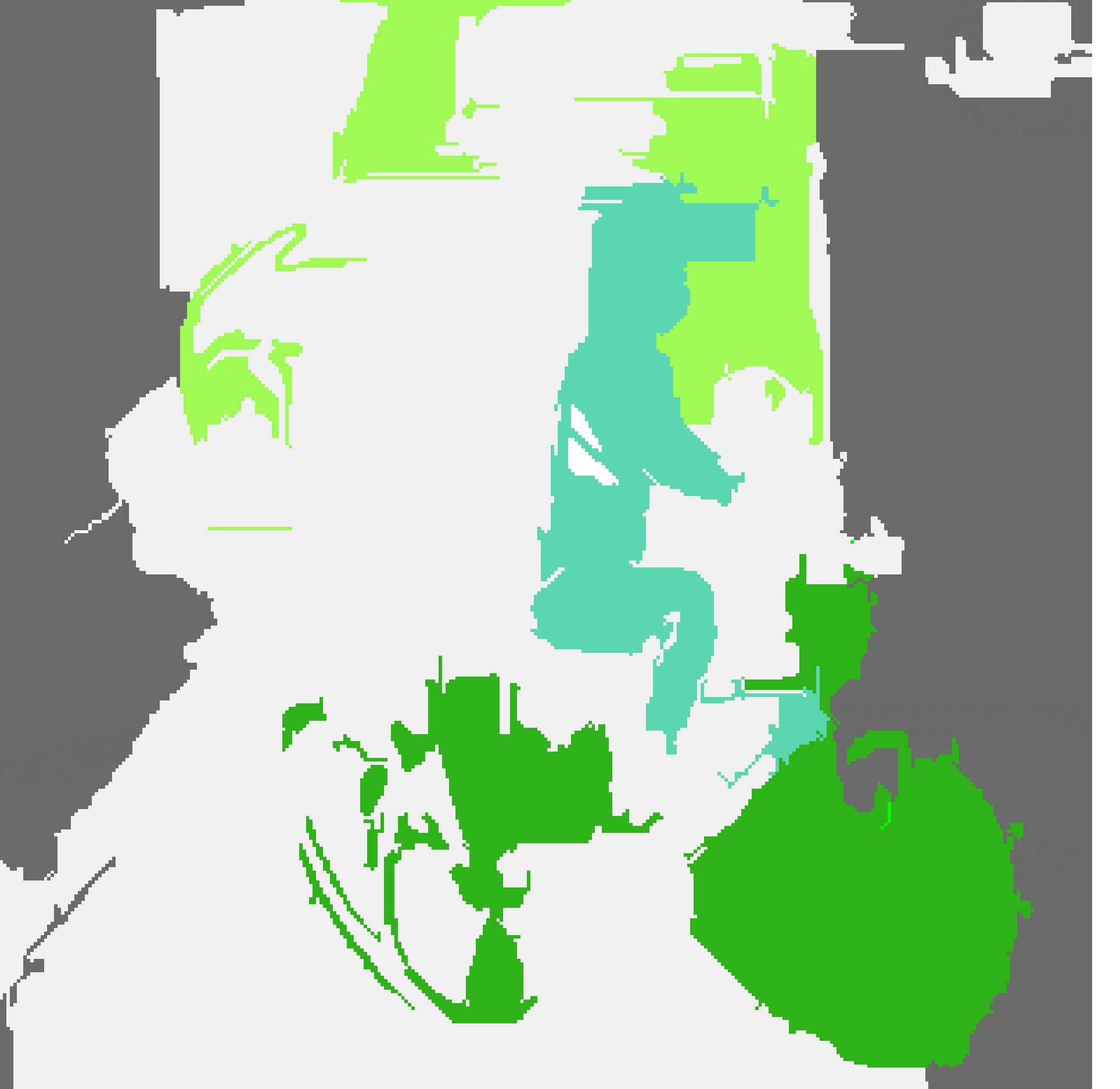}
     \scalebox{0.6}{Mixed Seed}
     \end{subfigure} 
     \caption{\small{The second feedback chain}}
     \label{fig:first_figure_feedback_two}
     \end{subfigure}
     
     \end{subfigure}
     
\setlength{\abovecaptionskip}{5pt} 	 
\setlength{\belowcaptionskip}{-15pt}

\caption{(a) Example of training image and its ground truth. (b) The pipeline of our closed-loop adaptation mechanism with two feedback chains. (c) Illustration of the dynamic seeds, generated by the first feedback chain, in several epochs of training. (d) The second feedback chain further expands the region of seeds by integrating a relationship matrix built upon superpixels. (The grey represents background and the white represents unlabeled/ignored pixels).}
\label{fig:structure_info}
\end{figure}

In this paper, we propose Dual-Feedback Network (DFN), a closed-loop system with two feedback chains, for weakly-supervised semantic segmentation. The architecture of DFN is shown in Figure \ref{fig:structure_info}. Our general idea is to adopt a divide-and-conquer strategy to recover both cross-image and inner-image structure information, and each feedback chain is imposed to compensate for the lack of one type of structure information. The first chain updates the labels progressively to correct errors made by static pseudo-label (see Figure~\ref{fig:structure_info}(c)). Specifically, we generate initial seeds using Class Activation Mapping (CAM)~\cite{zhou2016learning} and set it the pseudo-label to train the network in the first few epochs. After that, the network output is routed back to the initial seeds to amend their probability distributions. The first feedback chain repeats this procedure several iterations until seeds are not changed, whereas the second feedback chain uses a customized random walk to integrate inner-image structure information characterized by a relationship matrix built upon superpixels into network training (see Figure~\ref{fig:structure_info}(d)). Our method demonstrates state-of-the-art performance on the PASCAL VOC 2012 segmentation dataset.

To summarize, our contributions are threefold:
\begin{enumerate}
    \item 
    We interpret weakly-supervised semantic segmentation as a closed-loop problem, and introduce two feedback chains to recover, respectively, the cross-image structure information and the inner-image structure information. We also demonstrate that the new network, i.e., DFN, can be trained end-to-end.
    \item
    We construct a relationship matrix based on superpixels to recover inner-image structure information. This method is computational and memory efficient, and robust to noisy labels.
    \item 
    Our method outperforms existing weakly-supervised semantic segmentation methods with image-level tags. Especially, the mean Intersection-Over-Union (mIOU) values of our method are 60.0\% and 61.1\% on val and test sets, respectively. 
\end{enumerate}

\section{Related Work}
Compared with fully-supervised semantic segmentation, the main issue for weakly-supervised semantic segmentation is the lack of structure information that tells us how to organize the pixels. This problem becomes more serious if only image-level tags are available. Existing weakly-supervised semantic segmentation with image-level tags can be divided into two categories.

The first category uses a specialized loss function or objective to recover the structure information. \cite{pathak2014fully} initially suggests a multi-instance loss (MIL) function to recognize the most discriminatory region for each category. Later, \cite{pathak2015constrained} adds some linear constrains into an iteratively updating process to restrict the structure information of the output, leading to an improved segmentation result. On the other hand, \cite{durand2017wildcat:} proposes a weakly-supervised learning transfer layer to discover complementary regions for the subcategories which belong to the same category, whereas \cite{li2018tell} builds a guided attention inference network and uses an attention mining loss to constrain the output. However, this kind of methods only restricts a small part of pixels in the network output.  

Different from the first category, the methods in the second category generate pseudo-label to incorporate the structure information, and use the new generated labels to constrain all of the pixels in the network output. For example, \cite{kolesnikov2016seed} obtains the pseudo-label using CAM, \cite{saleh2016built-in} uses the activations from the middle layers of VGG-16~\cite{simonyan2015very} to generate some candidate regions, whereas \cite{hou2017bottom-up} sets attention map of each image as labels. \cite{wei2017stc:} proposes a simple-to-complex mechanism to generate the pseudo-label iteratively. The authors then apply an erasing mechanism into their network structure and discover new regions to expand the truth area in the pseudo-label \cite{wei2017object}. Despite substantial progress made by these methods, they always use static pseudo-label which is likely to accumulate errors incurred by initial inaccurate seeds or a dense conditional random field (DenseCRF) model~\cite{krahenbuhl2011efficient}  that is sensitive to noisy labels.

In this paper, we introduce two feedback chains into existing weakly-supervised semantic segmentation network to form a close-loop system, which enables iterative extraction of robust structure information and dynamical correction of errors made by inaccurate seed localization. Specifically, the first feedback chain updates the seed by the network output to recover the cross-image structure, whereas the second feedback chain incorporates robust inner-image structure information characterized by a relationship matrix built upon superpixels to further expand seed regions. It is worth noting that deep seed region growing (DSRG) \cite{huang2018weakly-supervised} also uses dynamic seed. However, it does not correct the inaccurate seed localization explicitly, which constrains the upper bound of segmentation performance. On the other hand, Mining Common Object Features (MCOF)~\cite{wang2018weakly-supervised} also suggests using superpixels to robustly rebuild inner-image structure information in the presence of noisy labels. Unlike MCOF, we use a simple relationship matrix to characterize superpixels, rather than an extra network that requires significantly higher computational and memory burden. Moreover, experiments demonstrate that our network outperforms both DSRG and MCOF with a large margin.

\section{Our Approach}
In this section, we elaborate our proposed DFN, especially the way to recover robust inner-image structure information with the aid of relationship matrix.

\subsection{Overview}
\begin{figure*}
\includegraphics[scale=0.5]{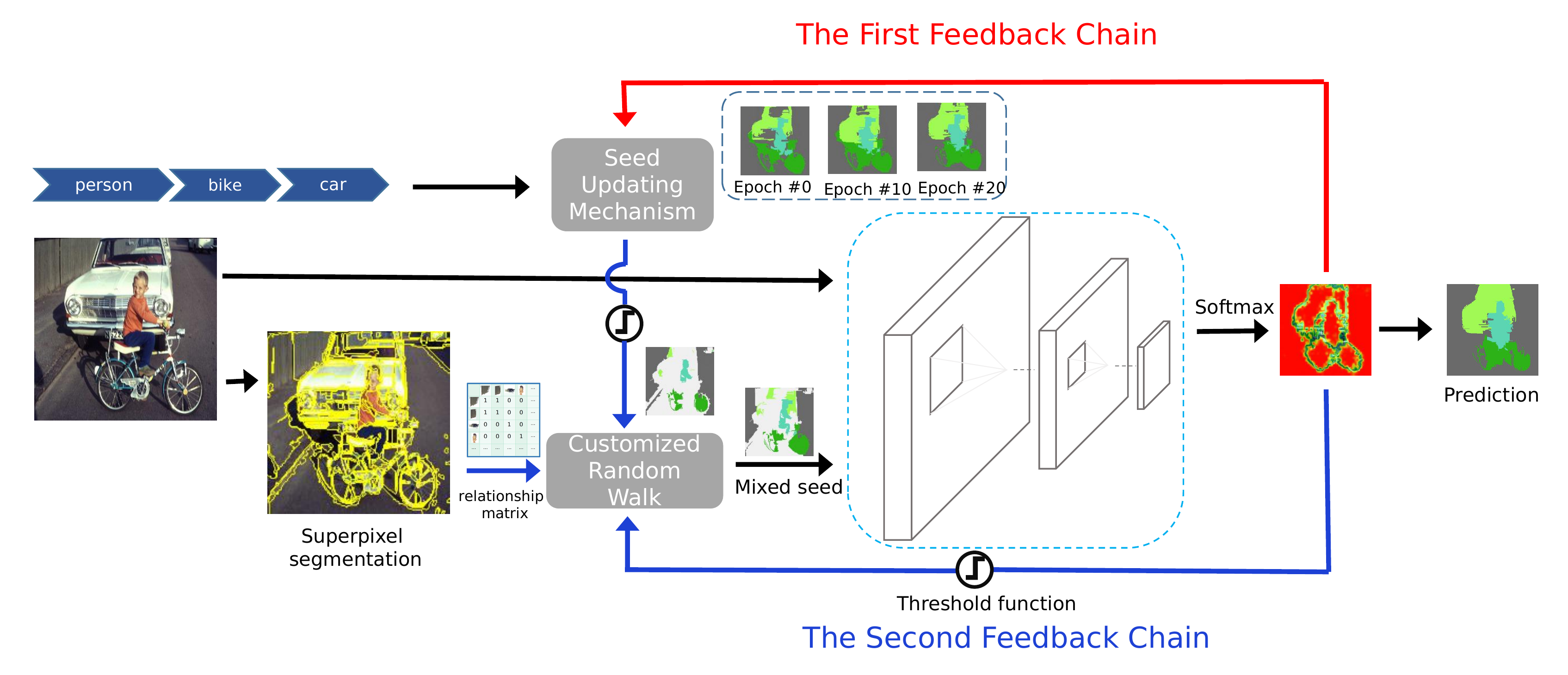}
\setlength{\abovecaptionskip}{1pt} 
\setlength{\belowcaptionskip}{-20pt}
\caption{A schematic illustration of our proposed Dual-Feedback Network with two feedback chains. The first feedback chain (red arrow) dynamically update the seeds with the aid of network output. The second feedback chain (blue arrows) merges the relationship matrix, the network output and the dynamic seeds using a customized random walk. The input to the network includes both the training image and the mixed seed.}
\label{fig:framework}
\end{figure*}

Similar to~\cite{kolesnikov2016seed,wang2018weakly-supervised,huang2018weakly-supervised}, we use DeepLab-LargeFOV network \cite{chen2018deeplab:} as the baseline network and build two feedback chains as shown in Figure~\ref{fig:framework}. Specifically, we first generate initial seeds using CAM and the saliency detection method in~\cite{jiang2013salient}. Then, in the second feedback chain, we integrate the robust inner-image structure information characterized by a relationship matrix built upon superpixels to produce the mixed seeds. Next, the baseline network is trained with the obtained mixed seeds under the seed loss function \cite{huang2018weakly-supervised} and the boundary loss function 
\cite{kolesnikov2016seed}. Using the network output, we update the seeds progressively along the first feedback chain. We repeat above procedures until there is no change on seeds.
\subsection{The First Feedback Chain}
Although the seed cues localized by CAM or saliency maps are discriminative and precise, they are sparsely distributed and still contain tiny errors.  To correct these errors, we adopt an updating mechanism to adjust seeds progressively. Specifically, in each iteration, we update the seed locations with the network output using the following simple formula:
\begin{equation}\label{eq_1}
    S_{new} = (1-w)\times S_{old} + w\times N_{out},
\end{equation}
where $S_{new}$ denotes seeds in the current iteration, $S_{old}$ denotes seeds from the last iteration, $N_{out}$ is the network output, and $w\in[0,1]$ is a weighting factor that balances the update rate.

\begin{figure*}
\includegraphics[scale=0.3]{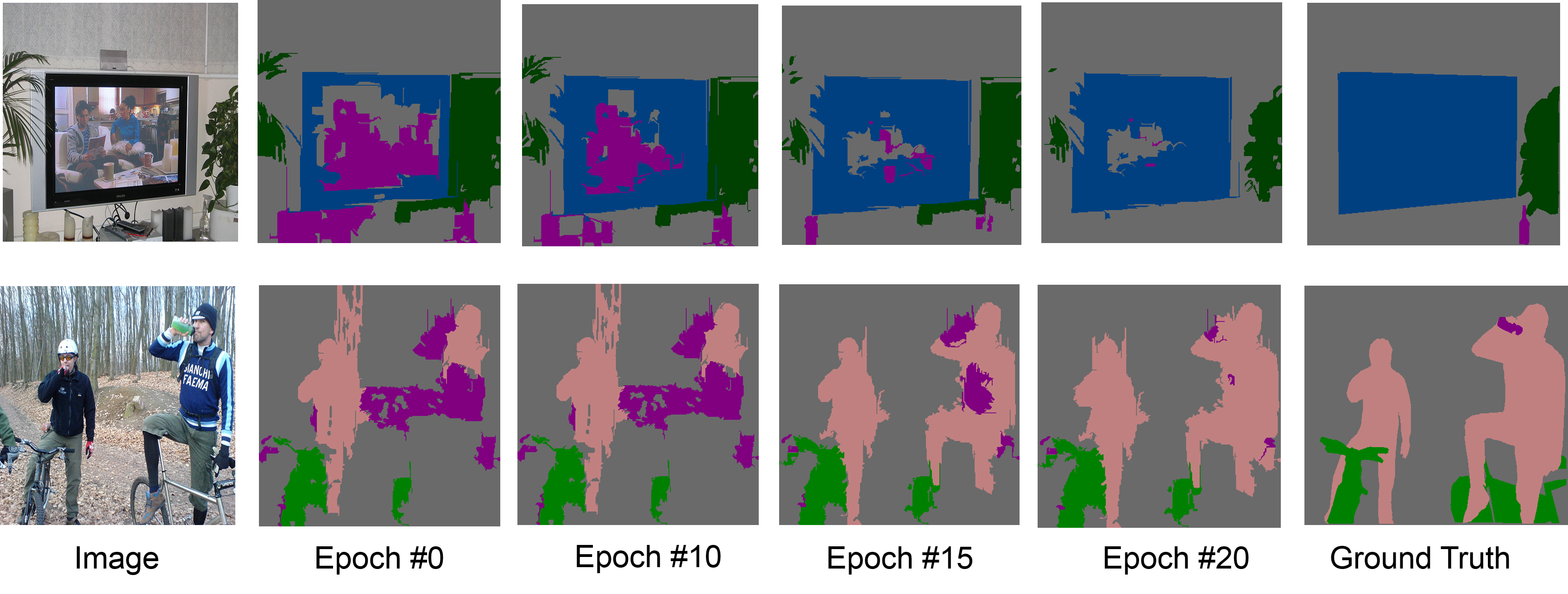}
\setlength{\abovecaptionskip}{3pt}
\setlength{\belowcaptionskip}{-15pt}

\caption{Results of the dynamic seeds, generated by the first feedback chain in several epochs during the training phase.}
\label{fig:first_feedback_chain}
\end{figure*}

One should note that there are two key points behind Eq.~(\ref{eq_1}). First, to preserve the boundary information and reduce the memory cost, the seeds are stored based on superpixels, rather than 41$\times$ 41 pixel blocks as in \cite{wei2017stc:} and \cite{huang2018weakly-supervised}. Second, in the interval of two subsequent seeds iterations, the network is trained independently with $3$ epochs. Moreover, we monitor the seeds states in different iterations to determine whether to continue the update. Specifically, each superpixel in seeds is characterized by a vector with its element corresponds to the probability to each category. Then, for one superpixel, we view it unchanged if the sum of absolute differences between elements of two vectors in two subsequent iterations is less than $0.1$. We stop seeds update if $95\%$ of superpixels are unchanged. Figure~\ref{fig:first_feedback_chain} shows the evolution of seeds at different epochs. As can be seen, with the increase of the number of iterations, initial errors are corrected and the dynamic seeds get closer to the ground truth.

\subsection{The Second Feedback Chain}
\subsubsection{Robust Inner-Image Structure Information Recovery}
\begin{figure}[h]
\centering
    \begin{subfigure}[b]{0.8\textwidth}
        \begin{subfigure}[b]{0.3\textwidth}
        \centering
        \includegraphics[width=\textwidth]{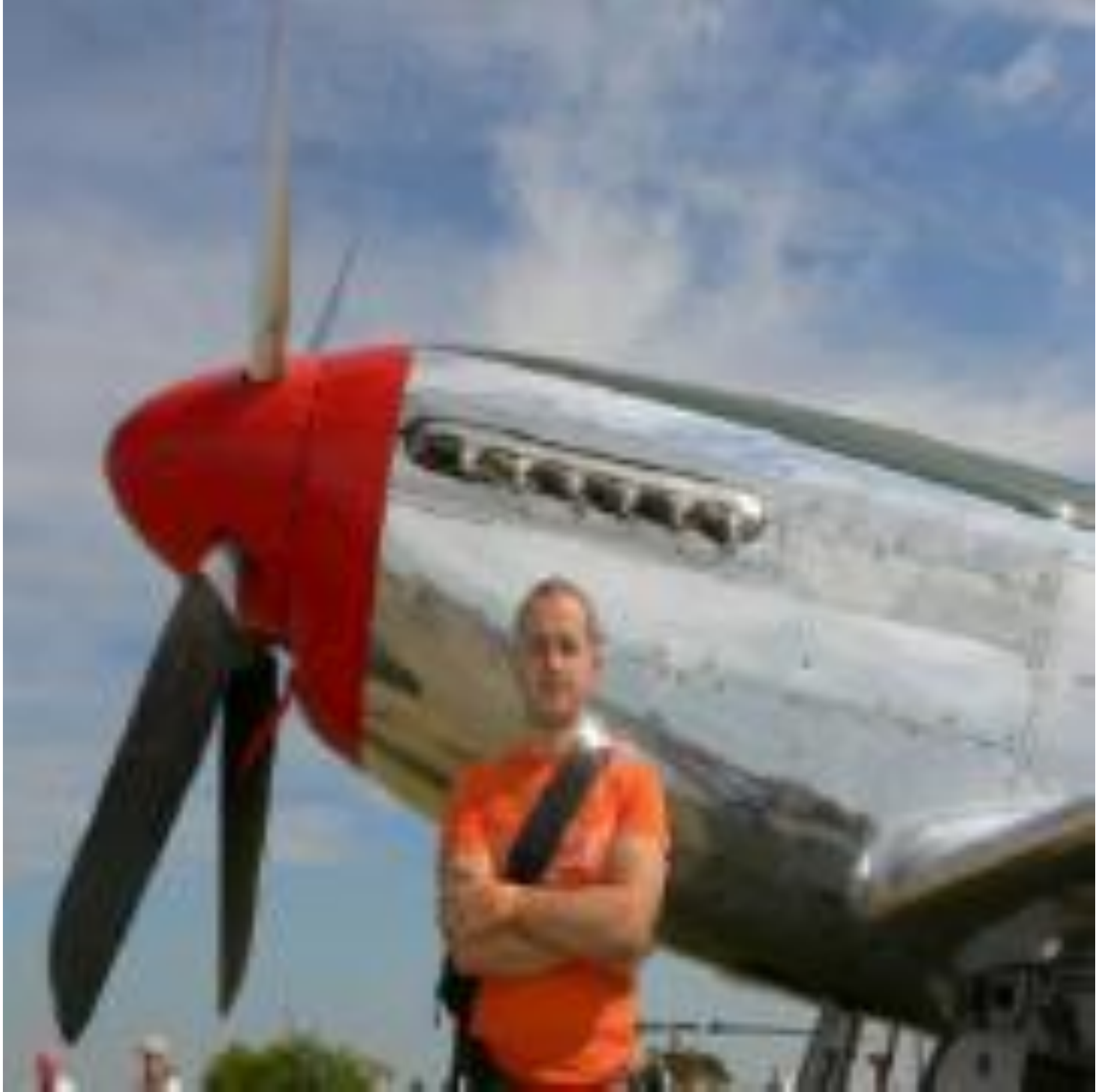}
        \small{(a) Training image}
        \end{subfigure}
        \hfill
        \begin{subfigure}[b]{0.3\textwidth}
        \centering
        \includegraphics[width=\textwidth]{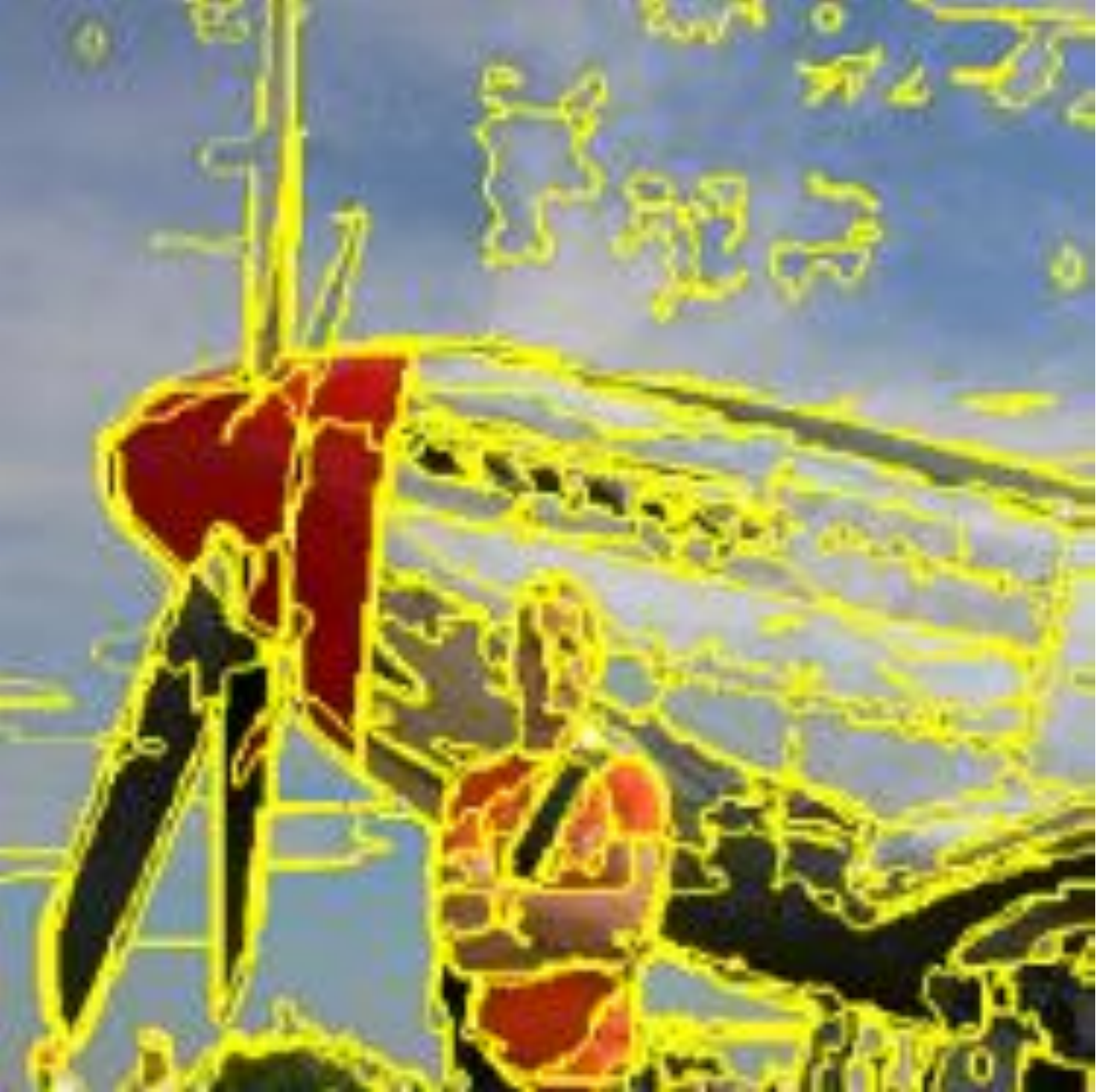}
        \footnotesize{(b) Result of Felzenszwalb}
        \end{subfigure}
        \hfill
        \begin{subfigure}[b]{0.3\textwidth}
        \centering
        \includegraphics[width=\textwidth]{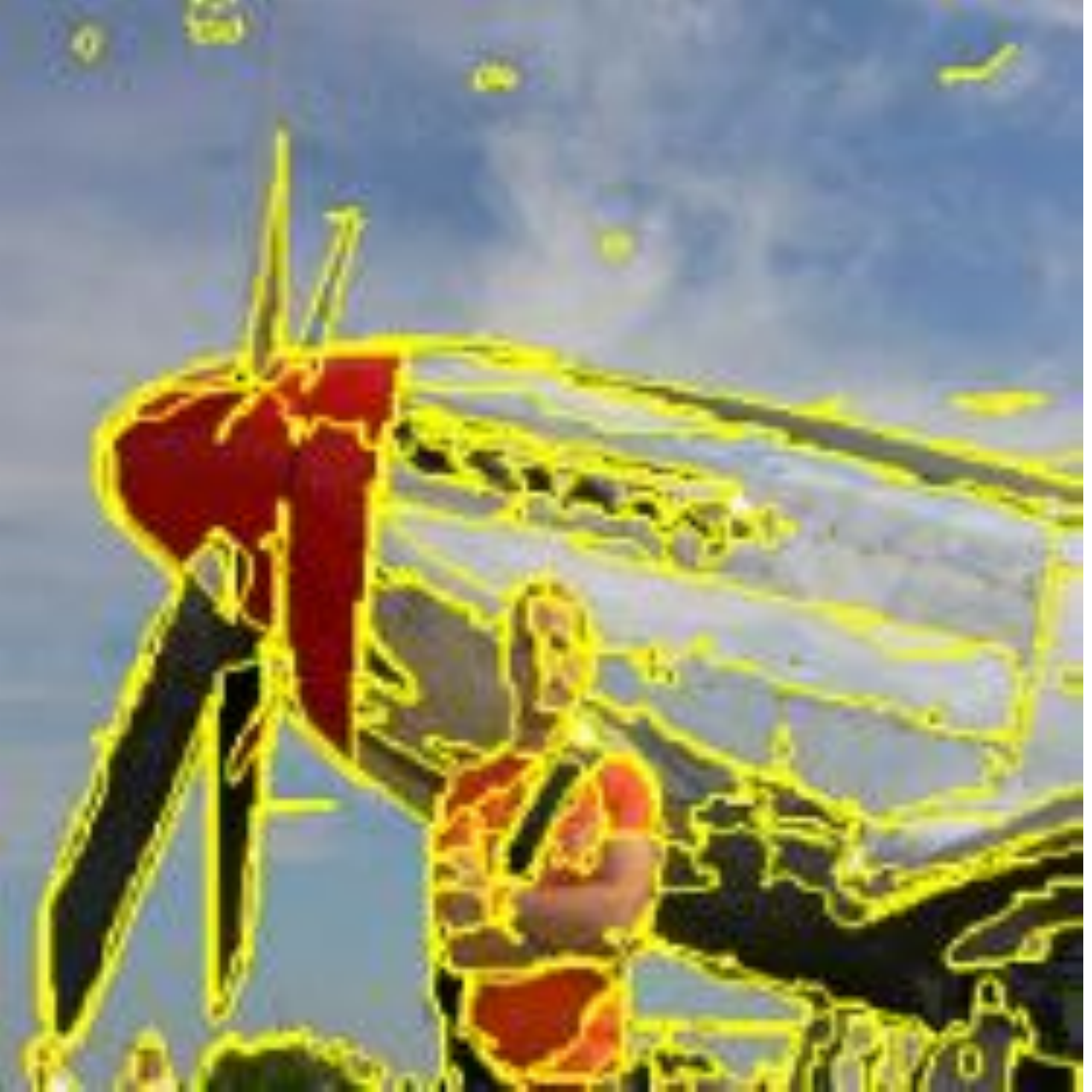}
        \small{(c) Our superpixels}
        \end{subfigure}
    \end{subfigure}
    
    \begin{subfigure}[b]{0.9\textwidth}

        \begin{subfigure}[b]{0.24\textwidth}
        \centering
        \includegraphics[width=\textwidth]{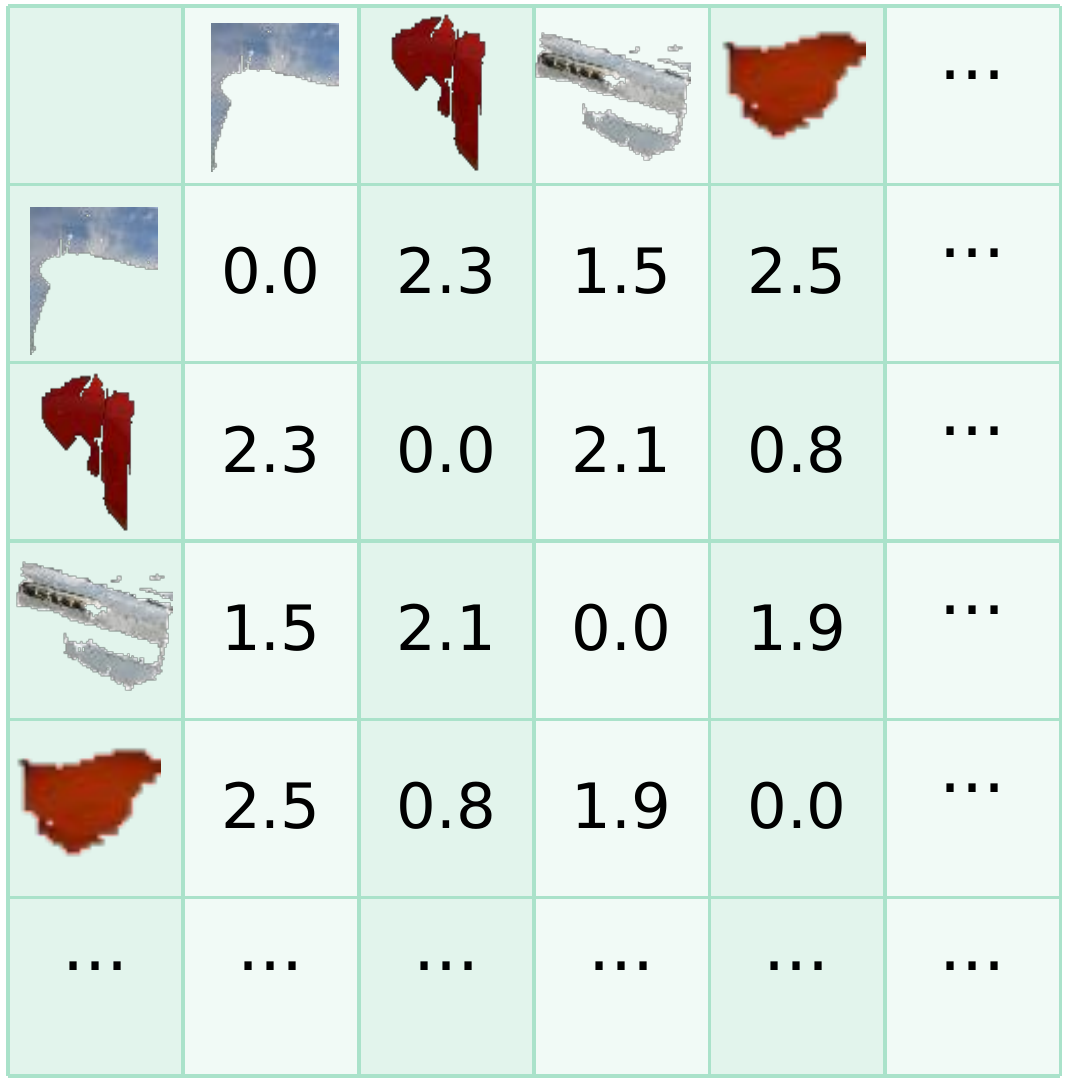}
        \footnotesize{(d) Distance matrix}
        \end{subfigure}
        \hfill
        \begin{subfigure}[b]{0.24\textwidth}
        \centering
        \includegraphics[width=\textwidth]{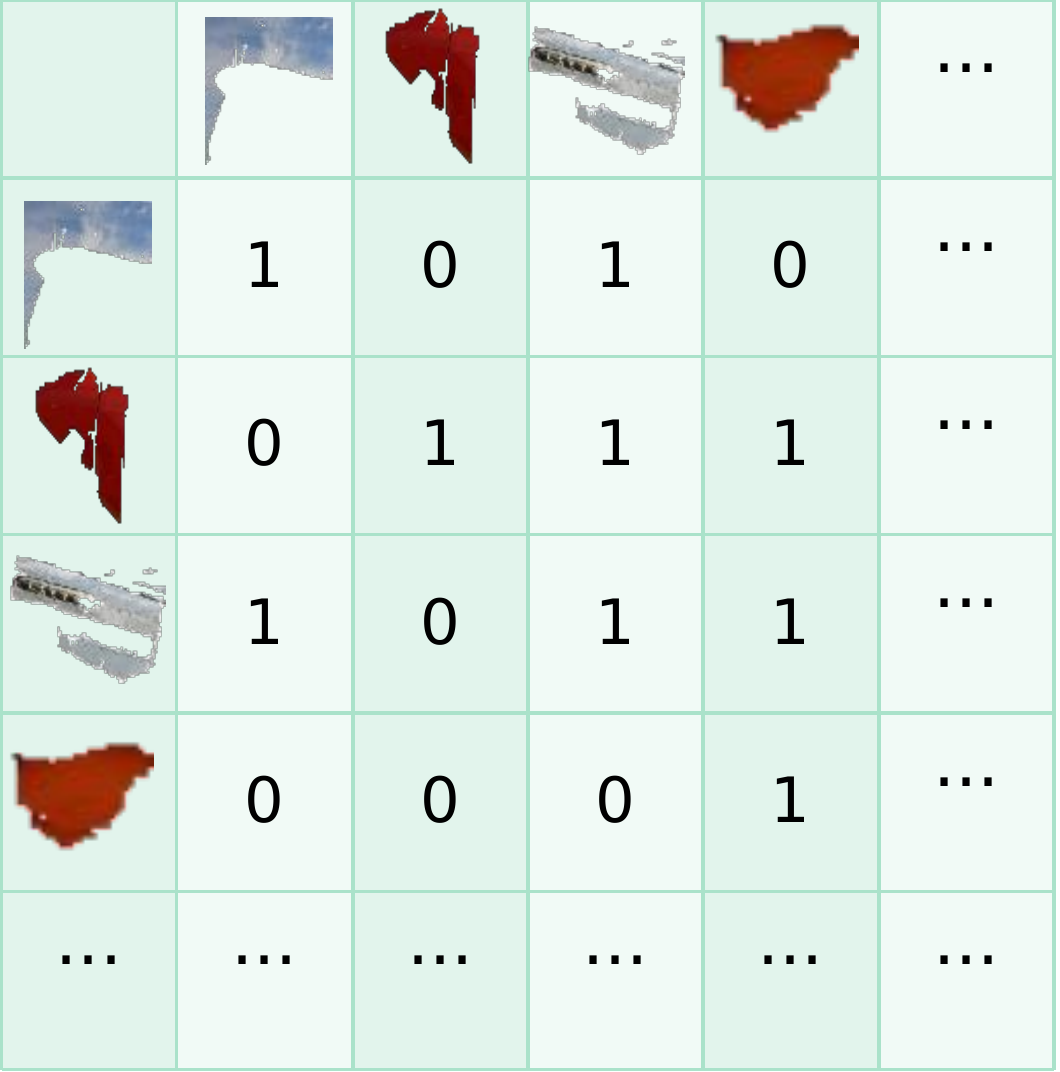}
        \footnotesize{(e) Similarity matrix}
        \end{subfigure}
        \hfill
        \begin{subfigure}[b]{0.24\textwidth}
        \centering
        \includegraphics[width=\textwidth]{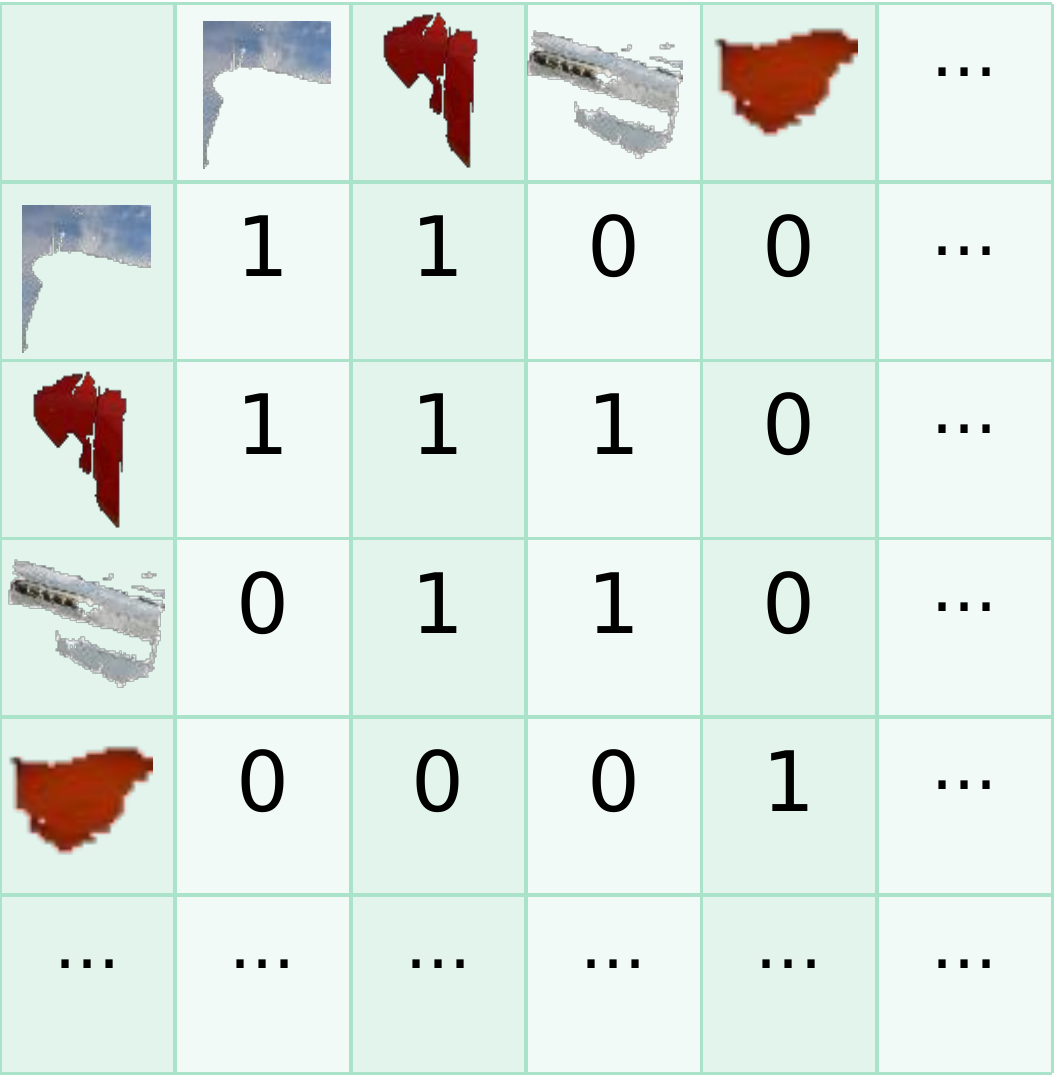}
        \footnotesize{(f) Adjacency matrix}
        \end{subfigure}
        \hfill
        \begin{subfigure}[b]{0.24\textwidth}
        \centering
        \includegraphics[width=\textwidth]{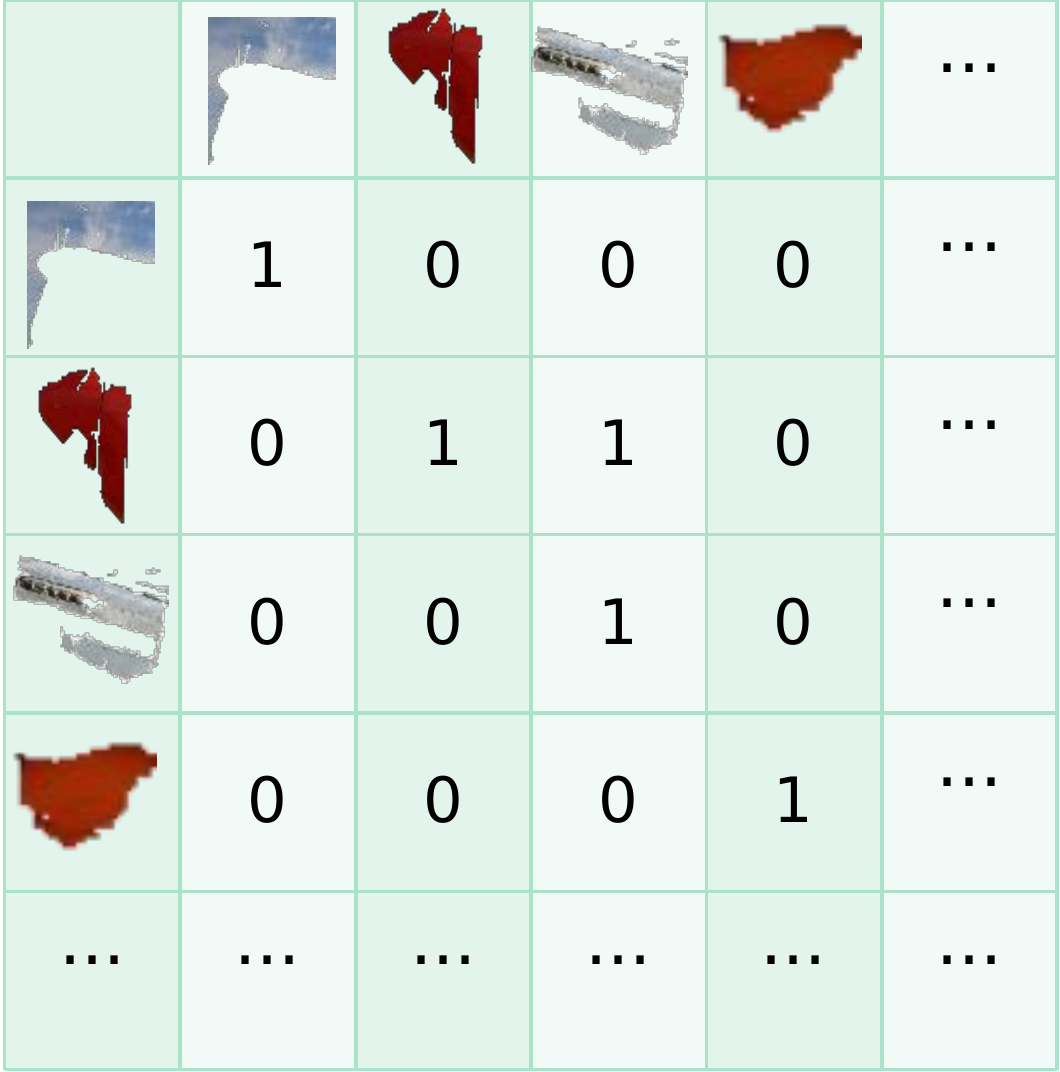}
        \scriptsize{(g) Relationship matrix}
        \end{subfigure}
    \end{subfigure}

    \caption{The top row orderly shows a training image, the superpixel segmentation result given by \cite{felzenszwalb2004efficient}, and our refined result. The bottom row shows the Euclidean distance matrix, the adjacency matrix, the distance matrix, and the relationship matrix respectively generated from superpixels.}
    \label{fig:process_for_inner_image_structure_info}
\end{figure}

Previous works use DenseCRF to build the relationships among pixels to recover the inner-image structure information. Unfortunately, the DenseCRF relies heavily on the color and location of each pixel, which makes it sensitive to noisy pixels. To address this limitation, we recover inner-image structrue information based on superpixels generated by Felzenszwalb's method \cite{felzenszwalb2004efficient} and refined by Region Adjacency Graph \cite{tremeau2000regions} (see the first row of Figure~\ref{fig:process_for_inner_image_structure_info} as an example).

The construction of relationship matrix consists of four steps. First, we generate Euclidean distance matrix for superpixels using zoom-out features~\cite{mostajabi2015feedforward}. However, unlike~\cite{mostajabi2015feedforward}, we only use the first two convolutional layers in VGG-16 to obtain low-level features. Nevertheless, owing to the diversity of objects under complex backgrounds in each image, the Euclidean distance with absolute values cannot precisely reflect the similarity between any two superpixels. Therefore, we then transform the Euclidean matrix to a simple similarity matrix (denoted $M_{siml}$), as shown in Figure~\ref{fig:process_for_inner_image_structure_info}(e). Specifically, we pick the $10$ smallest values in each row of the Euclidean matrix and set their values to $1$, indicating that they are mutually similar. We then set the values of remaining elements in each row of the Euclidean matrix to $0$, suggesting that they are not similar. Next, we construct an adjacency matrix (denoted $M_{adj}$) for superpixels to incorporate their location information. If two superpixels are neighboring, the corresponding element in the adjacency matrix is $1$, otherwise $0$ (see Figure~\ref{fig:process_for_inner_image_structure_info}(f)). We finally obtain a relationship matrix (denoted $M_{rel}$) with the following formula (see Figure \ref{fig:process_for_inner_image_structure_info}(f)):
\begin{equation}
    M_{rel} = M_{siml} \; \odot \; M_{adj},
\end{equation}
where $\odot $ denotes entry-wise product.

\subsubsection{Customized Random Walk}

After recovering the inner-image structure information characterized by a relationship matrix built upon superpixels, the second feedback chain performs a customized random walk to incorporate both the inner-image structure information and the cross-image structure information into network training. 

The input to our customized random walk includes relationship matrix, dynamic seeds, and network output. However, a preprocessing to dynamic seeds and network output is required herein. The reason is that each superpixel in seeds or network output is characterized by a vector with its element corresponds to the probability to each category. It means that if the value of the maximum element in the vector (of a superpixel) is small, we have a high classification uncertainty to assign this superpixel to its corresponding category. Moreover, the superpixels with high classification uncertainty will mislead the direction of seed expansion in random walk. Therefore, it is necessary to filter out these superpixels at first. To this end, we perform image thresholding to both dynamic seeds and network output. Specifically, we use a threshold function (denoted ${\rm Th_1}(\cdot)$) coupled with two hyperparameters $\alpha_{fg}$ and $\alpha_{bg}$ to distinguish foreground and background in seeds. Similarly, we use another threshold function (denoted ${\rm Th_2}(\cdot)$) coupled with hyperparameters $\beta_{fg}$ and $\beta_{bg}$ to distinguish foreground and background in the network output. For example, if the maximum element of one superpixel belong to the foreground in dynamic seeds is less than $\alpha_{fg}$, we drop out this superpixel in the following customized random walk.


By using the processed seeds as the initial state and the relationship matrix as the transition probability matrix, the basic random walk~\cite{Blum2018@foundations} generates candidate segmentation regions with ``$   {\rm Th_1}(S) \times M_{rel} $", where ``$\times$" denotes matrix multiplication. However, to remove false segmentation, we use the network output $N_{out}$ to guide the refinement of result from the basic random walk process. Therefore, the result $\hat S$ of our one-step customized random walk is given by:
\begin{equation}
\hat {S} = {\rm Th_1}(S) \times M_{rel} \odot {\rm Th_2}(N_{out}),
\end{equation}
where $\odot$ denotes entry-wise product. We repeat the operation of ``$\times M_{rel} \odot {\rm Th_2}(N_{out})$" $n$ times on the initial state ${\rm Th_1}(S)$ to obtain the $n$-step random walk result. For example, the result of two-step customized random walk is given by:
\begin{equation}
\hat {S} = {\rm Th_1}(S) \times M_{rel} \odot {\rm Th_2}(N_{out}) \times M_{rel} \odot {\rm Th_2}(N_{out}).
\end{equation}

\begin{figure}
\centering
    \begin{subfigure}[b]{1.1\textwidth}
        \begin{subfigure}[b]{0.45\textwidth}
            \begin{subfigure}[b]{0.3\textwidth}
            \centering
            \includegraphics[width=\textwidth]{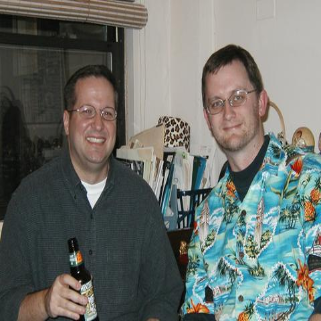}
            \end{subfigure}
            \hfill
            \begin{subfigure}[b]{0.3\textwidth}
            \centering
            \includegraphics[width=\textwidth]{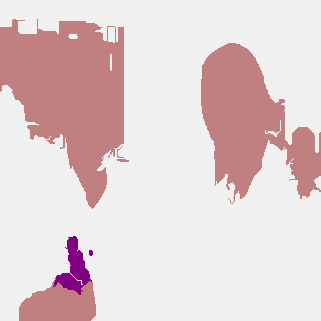}
            \end{subfigure}
            \hfill
            \begin{subfigure}[b]{0.3\textwidth}
            \centering
            \includegraphics[width=\textwidth]{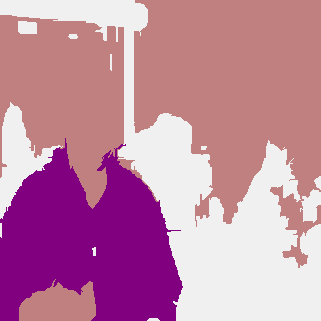}
            \end{subfigure}
        \end{subfigure}
        \hspace{0.003\textwidth}
        \begin{subfigure}[b]{0.45\textwidth}
            \begin{subfigure}[b]{0.3\textwidth}
            \centering
            \includegraphics[width=\textwidth]{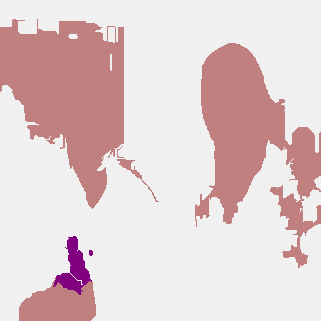}
            \end{subfigure}
            \hfill
            \begin{subfigure}[b]{0.3\textwidth}
            \centering
            \includegraphics[width=\textwidth]{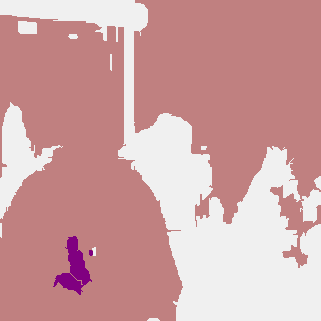}
            \end{subfigure}
            \hfill
            \begin{subfigure}[b]{0.3\textwidth}
            \centering
            \includegraphics[width=\textwidth]{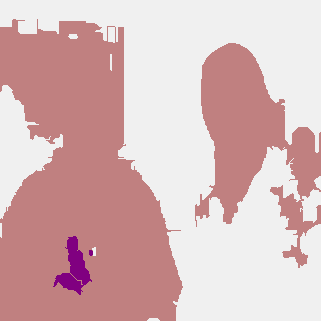}
            \end{subfigure}
        \end{subfigure}
    \end{subfigure}

    \begin{subfigure}[b]{1.1\textwidth}
                  \begin{subfigure}[b]{0.45\textwidth}  \begin{subfigure}[b]{0.3\textwidth}
            \centering
            \includegraphics[width=\textwidth]{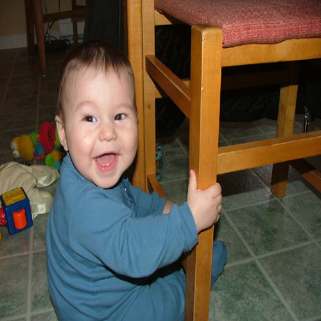}
            \scriptsize{(a)}
            \end{subfigure}
            \hfill
            \begin{subfigure}[b]{0.3\textwidth}
            \centering
            \includegraphics[width=\textwidth]{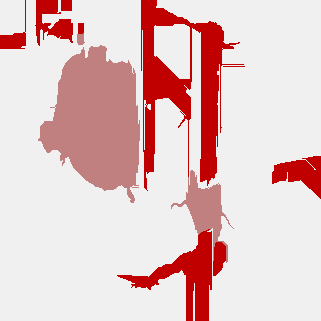}
            \scriptsize{(b)}
            \end{subfigure}
            \hfill
            \begin{subfigure}[b]{0.3\textwidth}
            \centering
            \includegraphics[width=\textwidth]{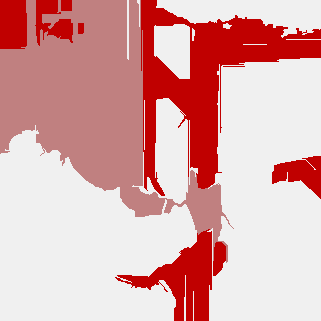}
            \scriptsize{(c)}
            \end{subfigure}

        \end{subfigure}
        \hspace{0.003\textwidth}
        \begin{subfigure}[b]{0.45\textwidth} \begin{subfigure}[b]{0.3\textwidth}
            \centering
            \includegraphics[width=\textwidth]{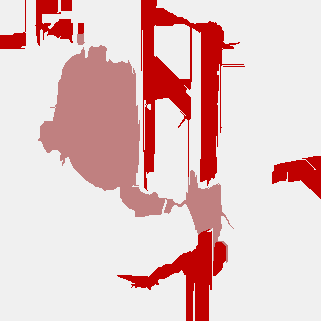}
            \scriptsize{(d)}
            \end{subfigure}
            \hfill
            \begin{subfigure}[b]{0.3\textwidth}
            \centering
            \includegraphics[width=\textwidth]{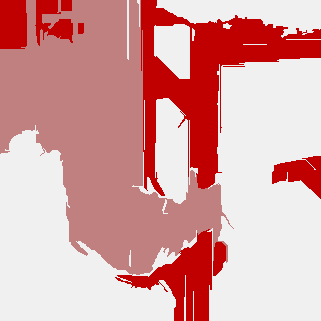}
            \scriptsize{(e)}
            \end{subfigure}
            \hfill
            \begin{subfigure}[b]{0.3\textwidth}
            \centering
            \includegraphics[width=\textwidth]{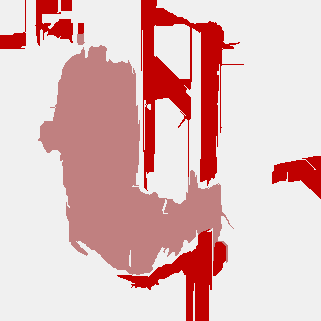}
            \scriptsize{(f)}
            \end{subfigure}
    \end{subfigure}
    \end{subfigure}

\caption{Evolution of seeds with our customized random walk: (a) training images; (b) initial object seeds; (c) mixed seeds with one-step random walk; (d) mixed seeds with one-step customized random walk; (e) mixed seeds with multi-step random walk; (f) mixed seeds with multi-step customized random walk.}
\label{fig:second_feedback_chain}
\end{figure}

We present two examples of our customized random walk in Figure~\ref{fig:second_feedback_chain}. By merging the robust inner-image structure information into the dynamic seeds, the obtained mixed seeds fit well with object boundaries and expand confident regions.

\section{Experiments}
\subsection{Experiment Setup}
\textbf{Dataset and Evaluation Metrics} We evaluate the proposed DFN on the PASCAL VOC 2012 segmentation benchmark dataset \cite{everingham2015the} which constains 20 foreground object classes and one background class. The segmentation part of VOC 2012 dataset is split into three parts: training (train, 1464 images), validation (val, 1449 images) and testing (test, 1456 images). Same to~\cite{huang2018weakly-supervised} and \cite{wang2018weakly-supervised}, the training set is extended with additional images from \cite{hariharan2011semantic}, resulting in an augmented set of 10582 images. Following the common practice, we use the mean Intersection-Over-Union (mIoU) criterion to compare our method with other approaches on both val and test sets. We report our results on standard val set with the ground truth segmentation masks are available.  For the test set, we submit the results of our final best model to the official evaluation server.

\textbf{Implementation details}
For all the experiments, we use the DeepLab-LargeFOV network~\cite{chen2018deeplab:} as the baseline segmentation architecture which is initialized from VGG-16~\cite{simonyan2015very} pre-trained on ImageNet \cite{deng2009imagenet:}. We use a mini-batch of $128$ images for SGD and initial learning rate of $3e$-$3$ which is decreased by a factor of $3$ every $5$ epochs. The momentum is $0.9$, the dropout rate is $0.5$ and the total epochs of training is $20$. 

For the first feedback chain, we set $w$ to $0.2$ and start to update the seeds every $3$ epochs after the $10$-th epoch. For the second feedback chain, we set $\alpha_{fg}$ to $0.90$, $\alpha_{bg}$ to $0.90$, $\beta_{fg}$ to $0.75$, and $\beta_{bg}$ to $0.90$. In addition, we perform two steps of customized random walk.
In the test phase, same to~\cite{kolesnikov2016seed,huang2018weakly-supervised}, the fully-connected CRF \cite{krahenbuhl2011efficient} and the multi-scale prediction \cite{chen2018deeplab:} are applied with their default parameters.

\textbf{Reproducibility} We use tensorflow to implement our approach. The code will be released soon. Anyone can use it and trains on a single NVIDIA GTX 1080TI GPU for about $15$ hours.

\subsection{Comparison with Previous Methods}
Table~\ref{table:val} and Table~\ref{table:test} report the mIoU values of our method on PASCAL VOC 2012 val and test sets, respectively, against previous state-of-the-art methods, namely EM-Adapt~\cite{papandreou2015weakly-}, STC~\cite{wei2017stc:}, SEC~\cite{kolesnikov2016seed}, MCOF~\cite{wang2018weakly-supervised}, and DSRG~\cite{huang2018weakly-supervised}. As can be seen, our method outperforms all compared methods in terms of mIoU value under the same experimental setting. Although MCOF~\cite{wang2018weakly-supervised} uses an extra network to recover structure information with superpixels, our method achieves a performance gain of $3.8\%$ and $3.5\%$ on val and test sets, respectively. Meanwhile, the improvement is $1.0\%$ and $0.7\%$ respectively compared to DSRG \cite{huang2018weakly-supervised}. 

\textbf{Computational and Memory Cost} We summarize the number of parameters and the number of floating-point operations (FLOPS) in Table~\ref{tab:params}. Compared with DSRG and MOCF, our network only requires approximately $50\%$ of parameters and reduces at least $30\%$ FLOPS. This result suggests that our method is more computational and memory efficient.

\textbf{Different Supervision Types} We compare our network with other methods under different types of supervisions. They are FCN \cite{long2015fully}, DeepLab \cite{chen2018deeplab:}, WSSL \cite{papandreou2015weakly-}, BoxSup \cite{dai2015boxsup:}, RAWK \cite{vernaza2017learning}, ScribbleSup \cite{lin2016scribblesup:}, and What'sPoint \cite{bearman2016what}. As can be seen in Table~\ref{tab:supervision_type}, our network achieves comparable performance to other methods that require stronger supervisions, e.g., the WSSL, the RWAK, or even the fully-supervised FCN. This result also suggests that there is a large performance gap between fully-supervised semantic segmentation and the weakly supervised semantic segmentation, especially for the points supervisions.
\begin{table*}
\tiny
\begin{tabular}{p{0.8cm}|p{0.30cm}p{0.30cm}p{0.30cm}p{0.30cm}p{0.30cm}p{0.30cm}p{0.30cm}p{0.30cm}p{0.30cm}p{0.30cm}p{0.30cm}p{0.30cm}p{0.30cm}p{0.30cm}p{0.40cm}p{0.40cm}p{0.30cm}p{0.30cm}p{0.30cm}p{0.30cm}p{0.3cm}|p{0.4cm}}
\hline
Method & bkg & plane & bike & bird & boat & bottle & bus & car & cat & chair & cow & table & dog & horse & motor & person & plant & sleep & sofa & train & tv & \textbf{mIoU} \\
\hline
\scalebox{0.75}{EM-Adapt} & 66.3 & 44.4 & 17.9 & 35.1 & 30.0 & 34.5 & 47.0 & 46.1 & 45.1 & 18.8 & 37.6 & 26.9 & 42.0 & 44.1 & 42.1 & 38.9 & 27.7 & 37.0 & 23.7 & 49.3 & 43.0 & 38.0  \\ 
STC & 84.5 & 68.0 & 19.5 & 60.5 & \textcolor{blue}{42.5} & 44.8 & 68.4 & 64.0 & 64.8 & 14.5 & 52.0 & 22.8 & 58.0 & 55.3 & 57.8 & 60.5 & \textcolor{blue}{40.6} & 56.7 & 23.0 & \textcolor{blue}{57.1} & 31.2 & 49.8 \\
SEC & 82.2 & 61.7 & 26.0 & 60.4 & 25.6 & 45.6 & 70.9 & 63.2 & 72.2 & 20.9 & 52.9 & \textcolor{red}{30.6} & 62.8 & 56.8 & 63.5 & 57.1 & 32.2 & 60.6 & 32.3 & 44.8 & 42.3 & 50.7 \\
MCOF & 85.8 & \textcolor{blue}{74.1} & 23.6 & 66.4 & 36.6 & \textcolor{blue}{62.0} & 75.5 & 68.5 & 78.2 & 18.8 & \textcolor{blue}{64.6} & \textcolor{blue}{29.6} & 72.5 & 61.6 & 63.1 & 55.5 & 37.7 & 65.8 & 32.4 & \textcolor{red}{68.4} & 39.9 & 56.2 \\
DSRG & \textcolor{blue}{87.5} & 73.1 & \textcolor{blue}{28.4} & \textcolor{blue}{75.4} & 39.5 & 54.5 & \textcolor{red}{78.2} & \textcolor{red}{71.3} & \textcolor{red}{80.6} & \textcolor{red}{25.0} & 63.3 & 25.4 & \textcolor{blue}{77.8} & \textcolor{blue}{65.4} & \textcolor{blue}{65.2} & \textcolor{red}{72.8} & \textcolor{red}{41.2} & \textcolor{blue}{74.3} & \textcolor{red}{34.1} & 52.1 & \textcolor{blue}{53.0} & \textcolor{blue}{59.0} \\
\textbf{Ours} & \textcolor{red}{88.0} & \textcolor{red}{76.3} & \textcolor{red}{33.7} & \textcolor{red}{79.0} & \textcolor{red}{48.7} & \textcolor{red}{62.5} & \textcolor{blue}{76.0} & \textcolor{blue}{69.0} & \textcolor{blue}{79.5} & \textcolor{blue}{23.4} &\textcolor{red}{66.3} & 13.7 & \textcolor{red}{78.1} & \textcolor{red}{69.8} & \textcolor{red}{67.3} & \textcolor{blue}{71.1} & 39.1 & \textcolor{red}{74.4} & \textcolor{blue}{33.6} & 51.7 & \textcolor{red}{58.8} & \textcolor{red}{60.0} \\
\hline
\end{tabular}

\setlength{\abovecaptionskip}{3pt}
\setlength{\belowcaptionskip}{-10pt}

\caption{Comparison of different weakly-supervised semantic segmentation methods on PASCAL VOC 2012 val set. The best two methods are marked with \textcolor{red}{red} and \textcolor{blue}{blue}, respectively.}
\label{table:val}
\end{table*}

\begin{figure}
\includegraphics[scale=0.45]{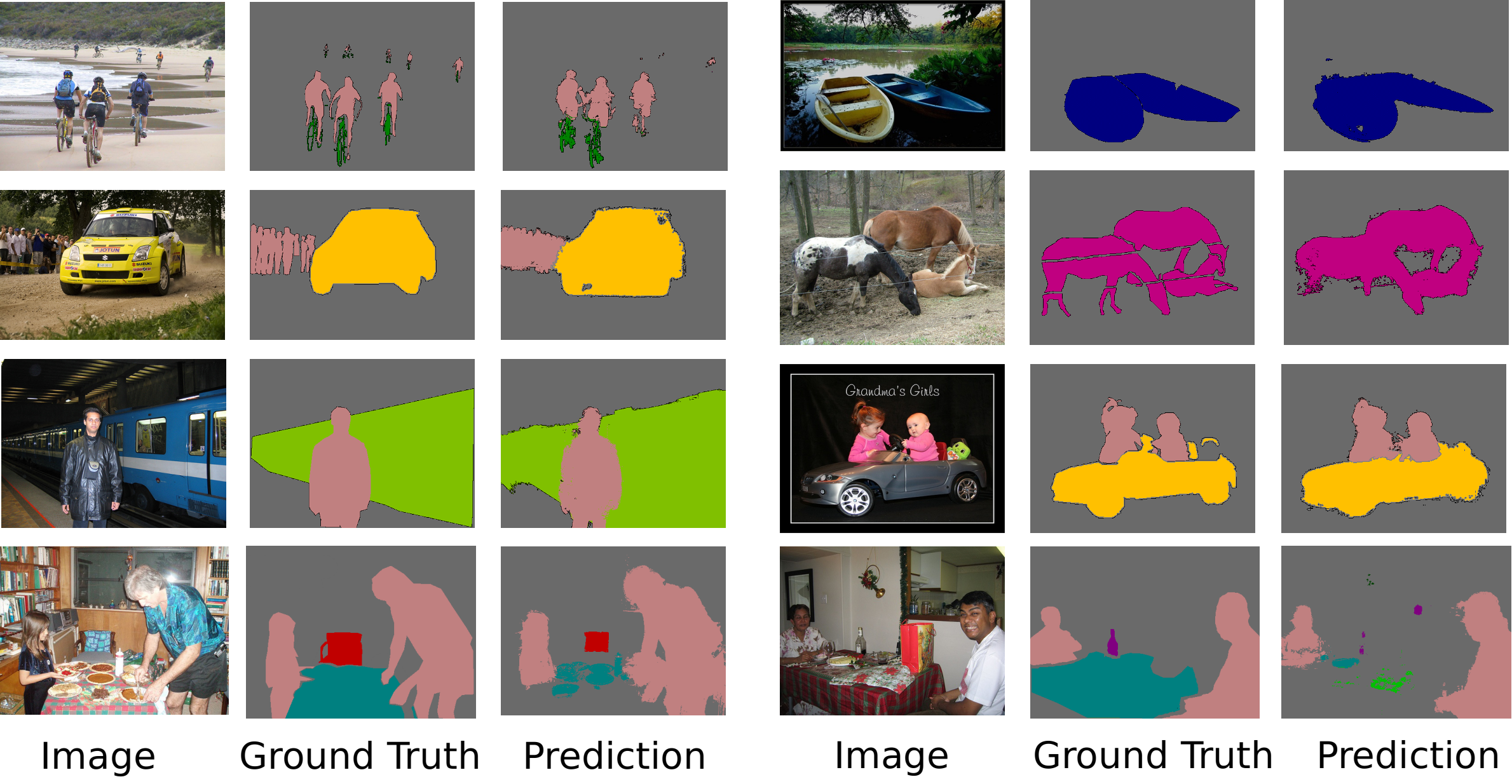}

\caption{Qualitative segmentation results on PASCAL VOC 2012 val set. One failure case is shown in the last row.}
\label{fig:examples}
\end{figure}

\begin{table*}
\tiny
\begin{tabular}{p{0.8cm}|p{0.30cm}p{0.30cm}p{0.30cm}p{0.30cm}p{0.30cm}p{0.30cm}p{0.30cm}p{0.30cm}p{0.30cm}p{0.30cm}p{0.30cm}p{0.30cm}p{0.30cm}p{0.30cm}p{0.40cm}p{0.40cm}p{0.30cm}p{0.30cm}p{0.30cm}p{0.30cm}p{0.3cm}|p{0.4cm}}

\hline
Method & bkg & plane & bike & bird & boat & bottle & bus & car & cat & chair & cow & table & dog & horse & motor & person & plant & sleep & sofa & train & tv & \textbf{mIoU} \\ \hline
\scalebox{0.75}{EM-Adapt}  & - & - & - & - & - & - & - & - & - & - & - & - & - & - & - & - & - & - & - & - & - & 39.6  \\ 
STC & 85.2 & 62.7 & 21.1 & 58.0 & 31.4 & 55.0 & 68.8 & 63.9 & 63.7 & 14.2 & 57.6 & 28.3 & 63.0 & 59.8 & 67.6 & 61.7 & 42.9 & 61.0 & 23.2 & \textcolor{blue}{52.4} & 33.1 & 51.2 \\
SEC & 83.5 & 56.4 & 28.5 & 64.1 & 23.6 & 46.5 & 70.6 & 58.5 & 71.3 & \textcolor{blue}{23.2} & 54.0 & 28.0 & 68.1 & 62.1 & 70.0 & 55.0 & 38.4 & 58.0 & 39.9 & 38.4 & 48.3 & 51.7 \\
MCOF & 86.8 & \textcolor{blue}{73.4} & 26.6 & 60.6 & 31.8 & 56.3 & \textcolor{blue}{76.0} & 68.9 & \textcolor{blue}{79.4} & 18.8 & 62.0 & \textcolor{red}{36.9} & 74.5 & 66.9 & 74.9 & 58.1 & 44.6 & 68.3 & 36.2 & \textcolor{red}{64.2} & 44.0 & 57.6 \\
DSRG & \textcolor{blue}{87.9} & 69.5 & \textcolor{red}{32.1} & \textcolor{blue}{74.2} & \textcolor{blue}{33.7} & \textcolor{blue}{59.4} & 74.9 & \textcolor{red}{71.5} & \textcolor{red}{80.1} & 21.9 & \textcolor{blue}{66.8} & \textcolor{blue}{32.7} & \textcolor{red}{76.4} & \textcolor{blue}{72.5} & \textcolor{red}{76.6} & \textcolor{blue}{73.4} & \textcolor{red}{49.9} & \textcolor{blue}{73.8} & \textcolor{blue}{43.4} & 42.0 & \textcolor{blue}{55.2} & \textcolor{blue}{60.4} \\
\textbf{Ours} & \textcolor{red}{88.0} & \textcolor{red}{73.9} & \textcolor{blue}{31.3} & \textcolor{red}{75.2} & \textcolor{red}{44.9} & \textcolor{red}{61.0} & \textcolor{red}{77.8} & \textcolor{blue}{69.6} & 78.9 & \textcolor{red}{24.9} & \textcolor{red}{71.9} & 20.2 & \textcolor{blue}{74.6} & \textcolor{red}{75.0} & \textcolor{blue}{75.3} & \textcolor{red}{74.3} & \textcolor{blue}{47.1} & \textcolor{red}{76.2} & \textcolor{red}{45.0} & 41.1 & \textcolor{red}{56.3} & \textcolor{red}{61.1} \\
\hline
\end{tabular}

\setlength{\abovecaptionskip}{5pt}
\setlength{\belowcaptionskip}{-18pt}

\caption{Comparison of different weakly-supervised semantic segmentation methods on PASCAL VOC 2012 test set. The best two methods are marked with \textcolor{red}{red} and \textcolor{blue}{blue}, respectively. The ``-" denotes unknown values that were not reported by the corresponding paper.}
\label{table:test}
\end{table*}
\textbf{Quantitative Results} The segmentation results shown in Figure~\ref{fig:examples} corroborate our quantitative evaluations. Note that, our method can generate precise segmentation results even for images containing complex backgrounds. However, our method is likely to fail when there are multiple small and dense objects on top of another larger object. Taking the last row of Figure \ref{fig:examples} as an example. Our method misclassified most of pixels in the table into background. One possible reason is that there are lots of plates and foods on the table, such that various of colors and textures of these small and dense objects make our method hard to generate a precise relationship matrix.


\begin{table}
    \centering
    \begin{tabular}{l|ccc}
    \hline 
         Method & Params~(M) & FLOPS~(G) & mIoU\\
    \hline 
         MCOF & about 40 & about 201 & 57.6 \\
 
        DSRG & 37.8 & 140.8 & 60.4 \\

        Ours & 20.6 & 101.9 & 61.1 \\
    \hline
    \end{tabular}
    
\setlength{\abovecaptionskip}{3pt}
\setlength{\belowcaptionskip}{-15pt}

    \caption{The number of parameters, the number of floating-point operations (FLOPS), and the mIoU values of our method, MCOF and DSRG.}
    \label{tab:params}
\end{table}
\begin{table}
    \centering
    \begin{tabular}{l|c|cc}
    \hline 
         Method & Supervision types & val & test \\
    \hline 
        FCN & fully-supervised & - & 62.2 \\
        
        DeepLab & fully-supervised & 67.6 & 70.3 \\
    \hline 
        WSSL & bounding-box & 60.6 & 62.2 \\

        BoxSup & bounding-box & 62.0 & 64.6 \\
    \hline
        RAWK & scribble & 61.4 & - \\

        ScribbleSup & scribble & 63.1 & - \\
    \hline
        What'sPoint & points & 46.0 & 43.6 \\ 
    \hline
        Ours & image-level tags & 60.0 & 61.1 \\
    \hline
    \end{tabular}
    
\setlength{\abovecaptionskip}{3pt}
\setlength{\belowcaptionskip}{-15pt}

    \caption{Comparison of semantic segmentation method under different supervision types on PASCAL VOC 2012 set.}
    \label{tab:supervision_type}
\end{table}

\subsection{Ablation Studies}
To validate the effects of different components, we perform some ablation experiments under different settings. In Table~\ref{tab:ablation}, we summarize the performance of our network in different degrading settings. Specifically, the ``baseline" indicates our baseline network without two feedback chains, the ``baseline+F1" indicates only integrating the first feedback chain into the baseline network, the ``baseline+F2" indicates only integrating the second feedback chain into the baseline network, whereas the ``baseline+F1+F2" indicates integrating both two chains. We observe $0.4$ percent gain in mIoU using the first feedback chain. This suggests that our dynamic seeds update mechanism indeed reduces more errors than DSRG. Moreover, by comparing the mIoU value of ``baseline+F2" with that of MCOF ($56.2$), although both methods attempt to recover inner-image structure information based on superpixels, our relationship matrix achieves $2.2$ percent performance gain, with significantly less computational and memory burden.

\begin{table}
    \centering
    \begin{tabular}{l|ccc}
    \hline 
         & accu & mIoU & fIoU \\
    \hline 
         baseline & 87.6 &  57.4 & 79.3 \\
    \hline   
     baseline~+~F1  & 88.0 & 57.8 & 80.0 \\
    \hline   
     baseline~+~F2  & 88.2 & 58.4 & 80.3 \\
    \hline    
     baseline~+~F1+~F2 & 89.3 & 60.0 & 81.4 \\
    \hline
    \end{tabular}
    
\setlength{\abovecaptionskip}{3pt}
\setlength{\belowcaptionskip}{-15pt}

    \caption{Comparison of our method under different settings on PASCAL VOC 2012 val set. The ``accu" denotes the pixel accuracy and the ``fIoU" denotes frequency weighted IoU.}
    \label{tab:ablation}
\end{table}

\subsection{Effects of Hyperparameters}
 We finally evaluate the performance of our network with respect to different hyperparameter settings and specify the way to pinpoint their values. The hyperparameter $w$ in Eq.~(\ref{eq_1}) represents the update rate in the first feedback chain. A large $w$ makes the dynamic seeds cannot supervise network training effectively, whereas a small $w$ slows down the correction of errors incurred by initial inaccurate seeds. We go through possible values of $w$ in a large range and select the best-performing one. From Table~\ref{tab:experiment_update_weight}, the optimal value of $w$ is $0.20$. 
 
\begin{figure}
\centering
\includegraphics[width=0.8\textwidth]{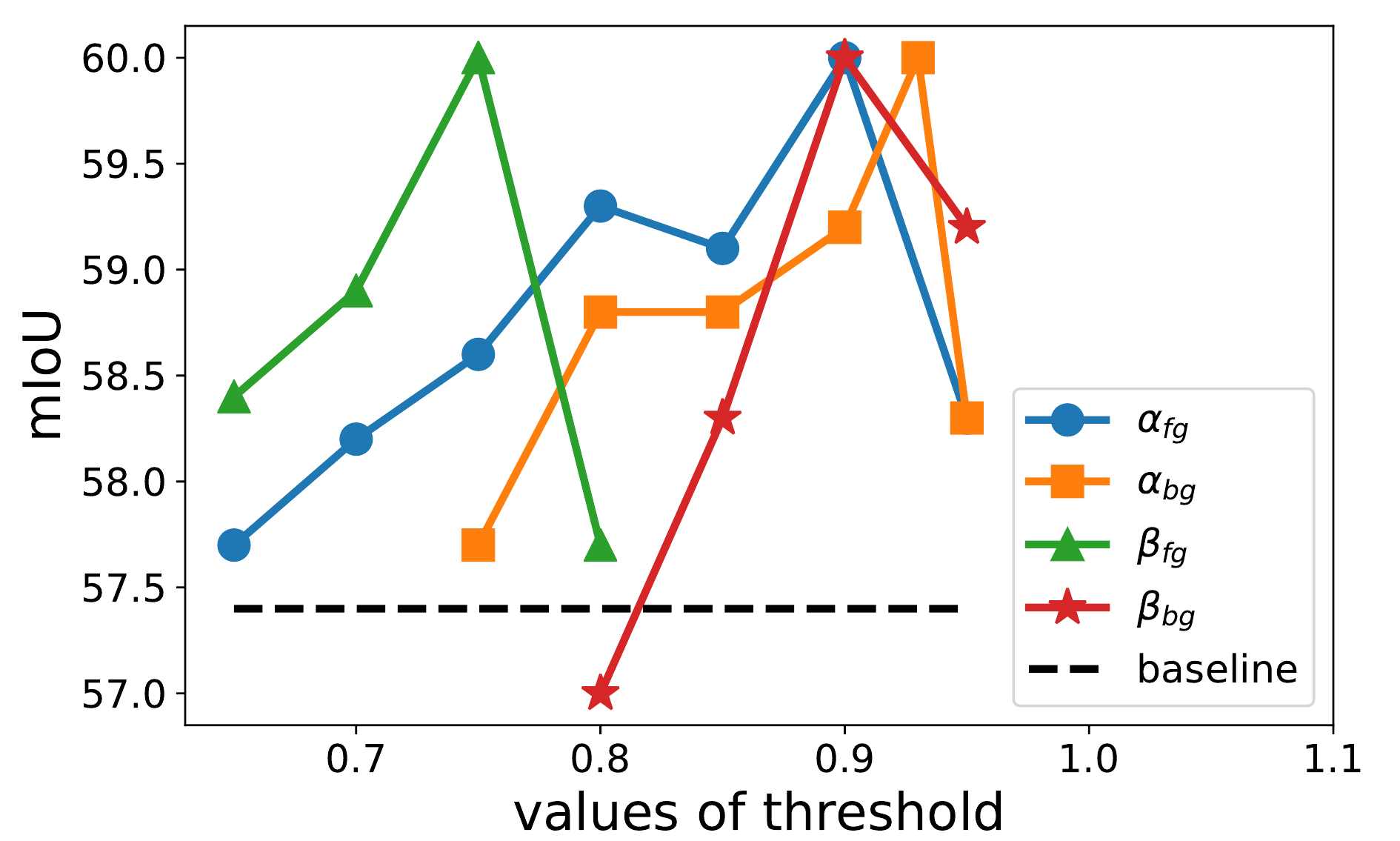}

\setlength{\abovecaptionskip}{2pt}
\setlength{\belowcaptionskip}{-15pt}

\caption{Performance of our method with respect to different values of $\alpha_{fg}$, $\alpha_{bg}$, $\beta_{fg}$ and $\beta_{bg}$. }
\label{fig:threshold}
\end{figure}

 In the second feedback chain, there are four hyperparameters, i.e., $\alpha_{fg}$, $\alpha_{bg}$, $\beta_{fg}$ and $\beta_{bg}$. We set their values to $\alpha_{fg}=0.90$, $\alpha_{bg}=0.90$, $\beta_{fg}=0.75$ and $\beta_{bg}=0.90$. These hyperparameters are tuned with a coarse-to-fine procedure. In the coarse module, we roughly determine a satisfactory range for each hyperparameter (for example, the range of $\alpha_{fg}$ is $[0.6,1.0]$). Then, in the fine-tuning module, we divide these parameters into two groups based on their correlations: (1) $\{\alpha_{fg},\alpha_{bg}\}$; (2) $\{\beta_{fg},\beta_{bg}\}$. When we test values of one group of hyperparameters, another group is set to default values, i.e., the mean value of the optimal range given by the coarse module. In each group, the hyperparameters are tuned with grid search (for example, $\alpha_{fg}$ is tuned at the range $[0.6,1.0]$ with an interval $0.1$). We finally pinpointed the specific value as the one that can achieve the highest mIoU value (for example, the final value of $\alpha_{fg}$ is $0.90$). We also test the sensitivity of these hyperparameters. To this end, we evaluate the performance variation for one hyperparameter with other three fixed. The results are shown in Figure~\ref{fig:threshold}. We can observe that in a wide range, our network outperforms the baseline with a large margin. It means that our performance is not sensitive to these four hyperparameters. Moreover, the performance variation for $\alpha_{fg}$ and $\alpha_{bg}$ seems less than that for $\beta_{fg}$ and $\beta_{bg}$. One possible reason is that seeds update slowly than network output.
 
\begin{table}
    \centering
    \begin{tabular}{c|ccc}
    \hline 
        $w$ & accu & mIoU & fIoU \\
    \hline   
         0.10 & 88.8 & 59.1 & 80.7 \\
    \hline    
         0.20 & 89.3 & 60.0 & 81.4 \\
    \hline    
         0.30 & 89.2 & 59.4 & 81.4 \\
    \hline
         0.40 & 89.0 & 58.7 & 80.9 \\
    \hline
    \end{tabular}

    \caption{Performance of our method for different $w$.}
    \label{tab:experiment_update_weight}
\end{table}

\section{Conclusion and future work}
In this paper, we designed a closed-loop network architecture, namely Dual-Feedback Network, for weakly-supervised semantic segmentation with image-level tags by introducing two feedback chains. The proposed network can correct the errors made by initial inaccurate seed localization with a seed updating mechanism and increase the robustness to noisy labels by incorporating inner-image structure information from superpixels. Experiments on PASCAL VOC 2012 show that our network outperforms previous state-of-the-art methods under the same experimental condition. Our method is also much more computational and memory efficient. Moreover, it is robust to different parameter settings.

In the future, we will continue improving the performance of DFN. This is because our prediction results in some examples still suffer from non-smooth boundaries or inconsistent small holes. At the same time, we are also interested in applying generative adversarial nets \cite{goodfellow2014generative} to recover the missing structure information.
\clearpage
\small
\bibliographystyle{named}
\bibliography{arXiv}

\begin{thebibliography}{}

\bibitem[\protect\citeauthoryear{Bearman \bgroup \em et al.\egroup
  }{2016}]{bearman2016what}
Amy Bearman, Olga Russakovsky, and Vittorio Ferrari, et~al.
\newblock What’s the point: Semantic segmentation with point supervision.
\newblock {\em ECCV}, pages 549--565, 2016.

\bibitem[\protect\citeauthoryear{Blum \bgroup \em et al.\egroup
  }{2018}]{Blum2018@foundations}
A.~Blum, J.~Hopcroft, and R.~Kannan.
\newblock Foundations of data science.
\newblock {\em [Online]. Available:https://www.cs.cornell.edu/jeh/book.pdf},
  2018.

\bibitem[\protect\citeauthoryear{Chen \bgroup \em et al.\egroup
  }{2018}]{chen2018deeplab:}
Liangchieh Chen, George Papandreou, and Iasonas Kokkinos, et~al.
\newblock Deeplab: Semantic image segmentation with deep convolutional nets,
  atrous convolution, and fully connected crfs.
\newblock {\em IEEE T-PAMI}, 40(4):834--848, 2018.

\bibitem[\protect\citeauthoryear{Dai \bgroup \em et al.\egroup
  }{2015}]{dai2015boxsup:}
Jifeng Dai, Kaiming He, and Jian Sun.
\newblock Boxsup: Exploiting bounding boxes to supervise convolutional networks
  for semantic segmentation.
\newblock {\em ICCV}, pages 1635--1643, 2015.

\bibitem[\protect\citeauthoryear{Deng \bgroup \em et al.\egroup
  }{2009}]{deng2009imagenet:}
Jia Deng, Wei Dong, and Richard Socher, et~al.
\newblock Imagenet: A large-scale hierarchical image database.
\newblock {\em CVPR}, pages 248--255, 2009.

\bibitem[\protect\citeauthoryear{Durand \bgroup \em et al.\egroup
  }{2017}]{durand2017wildcat:}
Thibaut Durand, Taylor Mordan, and Nicolas Thome, et~al.
\newblock Wildcat: Weakly supervised learning of deep convnets for image
  classification, pointwise localization and segmentation.
\newblock {\em CVPR}, pages 5957--5966, 2017.

\bibitem[\protect\citeauthoryear{Everingham \bgroup \em et al.\egroup
  }{2015}]{everingham2015the}
Mark Everingham, S~M Eslami, and Luc Van~Gool, et~al.
\newblock The pascal visual object classes challenge: A retrospective.
\newblock {\em IJCV}, 111(1):98--136, 2015.

\bibitem[\protect\citeauthoryear{Felzenszwalb and
  Huttenlocher}{2004}]{felzenszwalb2004efficient}
Pedro~F Felzenszwalb and Daniel~P Huttenlocher.
\newblock Efficient graph-based image segmentation.
\newblock {\em IJCV}, 59(2):167--181, 2004.

\bibitem[\protect\citeauthoryear{Goodfellow \bgroup \em et al.\egroup
  }{2014}]{goodfellow2014generative}
Ian~J Goodfellow, Jean Pougetabadie, and Mehdi Mirza, et~al.
\newblock Generative adversarial nets.
\newblock {\em NeurIPS}, pages 2672--2680, 2014.

\bibitem[\protect\citeauthoryear{Hariharan \bgroup \em et al.\egroup
  }{2011}]{hariharan2011semantic}
Bharath Hariharan, Pablo Arbelaez, and Lubomir~D Bourdev, et~al.
\newblock Semantic contours from inverse detectors.
\newblock {\em ICCV}, pages 991--998, 2011.

\bibitem[\protect\citeauthoryear{Hou \bgroup \em et al.\egroup
  }{2017}]{hou2017bottom-up}
Qibin Hou, Daniela Massiceti, and Puneet~Kumar Dokania, et~al.
\newblock Bottom-up top-down cues for weakly-supervised semantic segmentation.
\newblock {\em CVPR Workshop on Energy Minimization Methods}, pages 263--277,
  2017.

\bibitem[\protect\citeauthoryear{Huang \bgroup \em et al.\egroup
  }{2018}]{huang2018weakly-supervised}
Zilong Huang, Xinggang Wang, and Jiasi Wang, et~al.
\newblock Weakly-supervised semantic segmentation network with deep seeded
  region growing.
\newblock {\em CVPR}, pages 7014--7023, 2018.

\bibitem[\protect\citeauthoryear{Jiang \bgroup \em et al.\egroup
  }{2013}]{jiang2013salient}
Huaizu Jiang, Jingdong Wang, and Zejian Yuan, et~al.
\newblock Salient object detection: A discriminative regional feature
  integration approach.
\newblock {\em CVPR}, pages 2083--2090, 2013.

\bibitem[\protect\citeauthoryear{Kolesnikov and
  Lampert}{2016}]{kolesnikov2016seed}
Alexander Kolesnikov and Christoph~H Lampert.
\newblock Seed, expand and constrain: Three principles for weakly-supervised
  image segmentation.
\newblock {\em ECCV}, pages 695--711, 2016.

\bibitem[\protect\citeauthoryear{Krahenbuhl and
  Koltun}{2011}]{krahenbuhl2011efficient}
Philipp Krahenbuhl and Vladlen Koltun.
\newblock Efficient inference in fully connected crfs with gaussian edge
  potentials.
\newblock {\em NeurIPS}, pages 109--117, 2011.

\bibitem[\protect\citeauthoryear{Li \bgroup \em et al.\egroup
  }{2018}]{li2018tell}
Kunpeng Li, Ziyan Wu, and Kuanchuan Peng, et~al.
\newblock Tell me where to look: Guided attention inference network.
\newblock {\em CVPR}, pages 9215--9223, 2018.

\bibitem[\protect\citeauthoryear{Lin \bgroup \em et al.\egroup
  }{2016}]{lin2016scribblesup:}
Di~Lin, Jifeng Dai, and Jiaya Jia, et~al.
\newblock Scribblesup: Scribble-supervised convolutional networks for semantic
  segmentation.
\newblock {\em CVPR}, pages 3159--3167, 2016.

\bibitem[\protect\citeauthoryear{Long \bgroup \em et al.\egroup
  }{2015}]{long2015fully}
Jonathan Long, Evan Shelhamer, and Trevor Darrell.
\newblock Fully convolutional networks for semantic segmentation.
\newblock {\em CVPR}, pages 3431--3440, 2015.

\bibitem[\protect\citeauthoryear{Mostajabi \bgroup \em et al.\egroup
  }{2015}]{mostajabi2015feedforward}
Mohammadreza Mostajabi, Payman Yadollahpour, and Gregory Shakhnarovich.
\newblock Feedforward semantic segmentation with zoom-out features.
\newblock {\em CVPR}, pages 3376--3385, 2015.

\bibitem[\protect\citeauthoryear{Papandreou \bgroup \em et al.\egroup
  }{2015}]{papandreou2015weakly-}
George Papandreou, Liangchieh Chen, and Kevin~P Murphy, et~al.
\newblock Weakly- and semi-supervised learning of a dcnn for semantic image
  segmentation.
\newblock {\em arXiv}, 2015.

\bibitem[\protect\citeauthoryear{Pathak \bgroup \em et al.\egroup
  }{2014}]{pathak2014fully}
Deepak Pathak, Evan Shelhamer, and Jonathan Long, et~al.
\newblock Fully convolutional multi-class multiple instance learning.
\newblock {\em arXiv}, 2014.

\bibitem[\protect\citeauthoryear{Pathak \bgroup \em et al.\egroup
  }{2015}]{pathak2015constrained}
Deepak Pathak, Philipp Krahenbuhl, and Trevor Darrell.
\newblock Constrained convolutional neural networks for weakly supervised
  segmentation.
\newblock {\em ICCV}, pages 1796--1804, 2015.

\bibitem[\protect\citeauthoryear{Saleh \bgroup \em et al.\egroup
  }{2016}]{saleh2016built-in}
Fatemeh~Sadat Saleh, Mohammad~Sadegh Aliakbarian, and Mathieu Salzmann, et~al.
\newblock Built-in foreground/background prior for weakly-supervised semantic
  segmentation.
\newblock {\em ECCV}, 9912:413--432, 2016.

\bibitem[\protect\citeauthoryear{Simonyan and
  Zisserman}{2015}]{simonyan2015very}
Karen Simonyan and Andrew Zisserman.
\newblock Very deep convolutional networks for large-scale image recognition.
\newblock {\em ICIR}, 2015.

\bibitem[\protect\citeauthoryear{Tremeau and
  Colantoni}{2000}]{tremeau2000regions}
Alain Tremeau and Philippe Colantoni.
\newblock Regions adjacency graph applied to color image segmentation.
\newblock {\em IEEE T-IP}, 9(4):735--744, 2000.

\bibitem[\protect\citeauthoryear{Vernaza and
  Chandraker}{2017}]{vernaza2017learning}
Paul Vernaza and Manmohan Chandraker.
\newblock Learning random-walk label propagation for weakly-supervised semantic
  segmentation.
\newblock {\em CVPR}, pages 2953--2961, 2017.

\bibitem[\protect\citeauthoryear{Wang \bgroup \em et al.\egroup
  }{2018}]{wang2018weakly-supervised}
Xiang Wang, Shaodi You, and Xi~Li, et~al.
\newblock Weakly-supervised semantic segmentation by iteratively mining common
  object features.
\newblock {\em CVPR}, pages 1354--1362, 2018.

\bibitem[\protect\citeauthoryear{Wei \bgroup \em et al.\egroup
  }{2017a}]{wei2017object}
Yunchao Wei, Jiashi Feng, and Xiaodan Liang, et~al.
\newblock Object region mining with adversarial erasing: A simple
  classification to semantic segmentation approach.
\newblock {\em CVPR}, pages 6488--6496, 2017.

\bibitem[\protect\citeauthoryear{Wei \bgroup \em et al.\egroup
  }{2017b}]{wei2017stc:}
Yunchao Wei, Xiaodan Liang, and Yunpeng Chen, et~al.
\newblock Stc: A simple to complex framework for weakly-supervised semantic
  segmentation.
\newblock {\em IEEE T-PAMI}, 39(11):2314--2320, 2017.

\bibitem[\protect\citeauthoryear{Zhao \bgroup \em et al.\egroup
  }{2017}]{zhao2017pyramid}
Hengshuang Zhao, Jianping Shi, and Xiaojuan Qi, et~al.
\newblock Pyramid scene parsing network.
\newblock {\em CVPR}, pages 6230--6239, 2017.

\bibitem[\protect\citeauthoryear{Zhou \bgroup \em et al.\egroup
  }{2016}]{zhou2016learning}
Bolei Zhou, Aditya Khosla, and Agata Lapedriza, et~al.
\newblock Learning deep features for discriminative localization.
\newblock {\em CVPR}, pages 2921--2929, 2016.

\end{thebibliography}







\end{document}